\RequirePackage{amsmath}
\RequirePackage{fix-cm}
\documentclass[smallextended]{svjour3}       
\smartqed  
\usepackage{graphicx}
\usepackage{csvsimple}
\usepackage{datatool}
\usepackage{geometry}
\usepackage{booktabs}
\usepackage{varwidth}
\usepackage{array}
\usepackage[table]{xcolor}
\usepackage{tabularx}
\usepackage[ruled,vlined, algosection, procnumbered]{algorithm2e}

\arrayrulecolor{black!30}
\newcolumntype{?}{!{\color[gray]{0.7}\vrule width .4pt}}

\usepackage[backend=biber,style=authoryear,bibencoding=utf8]{biblatex}
\begin{document}

\title{Emergent and Unspecified Behaviors in Streaming Decision Trees
}
\subtitle{Why Online Decision Trees Perform So Well}


\author{Chaitanya Manapragada \and
        Geoffrey I Webb \and 
        Mahsa Salehi \and
        Albert Bifet
}


\institute{Chaitanya Manapragada, Geoff Webb, Mahsa Salehi \at
              Monash University, Australia \\
              \email{FirstName.LastName@monash.edu}           
           \and
           Albert Bifet \at
              University of Waikato, New Zealand \\
              \email{abifet@waikato.ac.nz}              
}

\date{Received: date / Accepted: date}

\maketitle

\begin{abstract}

\keywords{Concept Drift \and Hoeffding Tree \and Explainability}

Hoeffding trees are the state-of-the-art methods in decision tree
learning for evolving data streams. These very fast decision trees are
used in many real applications where data is created in real-time due
to their efficiency. In this work, we extricate explanations for why
these streaming decision tree algorithms for stationary and
nonstationary streams (HoeffdingTree and HoeffdingAdaptiveTree) work
as well as they do. In doing so, we identify thirteen unique
unspecified design decisions in both the theoretical constructs and
their implementations with substantial and consequential effects on
predictive accuracy—design decisions that, without necessarily
changing the essence of the algorithms, drive algorithm performance.
We begin a larger conversation about explainability not just of the
model but also of the processes responsible for an algorithm’s
success.

\end{abstract}

\section{Introduction}
\label{intro}

Algorithm design is, for now, mostly a manual, creative process; proposed strategies usually have major components assumed to cause the bulk of a behavior. But are we certain of our understanding of the causal relationships underpinning algorithmic performance? 

A learning system may consist of many components, all interacting in complex ways with existing mechanisms and leading to a system that is difficult to comprehend. Further, implementation details are quite often open to interpretation---it would be extremely restricting to formally specify all aspects of implementation. Given a published algorithm, unspecified differences in implementation decisions may lead to---overall---hugely differing behavior and may be best termed \textit{unspecified strategies}. 

Minor aspects of theory and implementation may be considered  irrelevant to the essence of the algorithm but may silently exert tremendous influence on measured objectives. For instance, we show that though Hoeffding Tree constitutes a major advance in incremental learning due to the introduction of the Hoeffding Test, a minor detail in how node statistics are initialized upon node creation has a significant role to play in its performance. This ``minor detail" pertaining the initialization of node statistics is not considered a key aspect of the algorithm. It is merely a convenience. But it is a convenience that has a strikingly large impact on the performance of HoeffdingTree, as we demonstrate in Section \ref{ht}.

The complexity of interactions between components of a learning system is such that the most unassuming of algorithm components may interact with other parts of the system to produce unexplained variations of performance. While these are subsumed into the overall aesthetic and direction of the algorithm developer and the algorithm itself, thus resisting straightforward detection, they warrant study and explanation, especially so when effects are unexpectedly significant. It is with this rationale that we study the benchmark MOA \parencite{bifet2010moa} implementations of two state-of-the-art stream learning methods: Hoeffding Tree \parencite{domingos2000mining}, and Hoeffding Adaptive Tree \parencite{bifet2009adaptive}. We find that simple design decisions in the algorithms and their respective implementations that are well within the scope of reasonable interpretation lead to substantial and consequential variations in prequential accuracy.

The main contributions of this paper are the following:
\begin{enumerate}

	\item Identification of unspecified design decisions in Hoeffding Tree and Hoeffding Adaptive Tree

	\item A thorough experimental study of how such unspecified features, individually and in combination, influence algorithm behavior and prequential error performance

	\item Suggestions for unspecified features that may be worth formally inducting into the algorithms

	\item A discriminative testbench that helped us identify differences in behavior and prequential error performance

\end{enumerate}

The outline of this paper is as follows: Section \ref{background} presents Background and Related Work; Section \ref{ht} details unspecified features and emergent behaviors due to them in the MOA implementation of Hoeffding Tree; Section \ref{hat} details unspecified features and emergent behaviors due to them in the MOA implementation of Hoeffding Adaptive Tree; and finally Section \ref{conclusions} contains our Conclusions.

\section{Background and Related Work}
\label{background}

\subsection{Batch learning origins of Decision Trees}

Decision trees were some of the earliest mechanisms identified \parencite{hunt1962concept} as highly interpretable models for the \textit{representation and storage} of knowledge. This representation of knowledge was initially in the form of a simple binary tree. All of the data were collected at the ``root" vertex of the tree; then, the datapoints were separated into the two child vertices (``nodes") based on a decision pertaining to the value of each datapoint. For instance, with unidimensional data $(x_1, x_2...x_m)$ obtained from the real line observed at the root node $N_{root}$, the data point $x_i$ might assigned to the child node on the left $N_{Left}$ if less than 0, and to the node on the right $N_{Right}$ if greater than or equal to zero. Further subdivision may be possible in recursive manner. This binary data structure is easily generalised---the data may also be multivariate and nominal (e.g. $\vec{x_i} = (Sunny, Cold, Windy)$), and trees are not required to be binary (for instance they may be ternary---with the root having child nodes $N_{Right}, N_{Middle}, N_{Left}$).

Algorithms for the construction of decision trees were consequently proposed; major milestones begin in 1966 with the Concept Learning System by Hunt et al. \parencite{earl1966experiments}, followed on by Quinlan in 1979 with ID3 \parencite{quinlan1979discovering, quinlan1983learning, Quinlan86inductionof}, Breiman in 1984 with Classification and Regression Trees (CART) \parencite{breiman1984classification}, and C4.5 by Quinlan in 1992 \parencite{Quinlan:2031749}.

These algorithms share a common core instrumentation; a split criterion determines how each node, starting with the root, splits data that have filtered to it; child nodes are created, and the tree is thus recursively grown. Early systems assumed data could be perfectly separated, and that data and trees were both binary; however, CART \parencite{breiman1984classification} and C4.5 \parencite{Quinlan:2031749} were robust systems that did not make assumptions of perfect data separability, pruned themselves to avoid overfitting, allowed multivariate and nominal data, and thus were suitable for a wide range of applications.

C4.5 by Ross Quinlan has remained a widely used decision tree algorithm for batch learning---a learning paradigm in which all of the data is available at once. It uses Information Gain as the heuristic for deciding best splits. Information Gain tells us what the relative class purity of two given class distributions is (the idea is that  ``purer" class distributions contain less information---one needs a shorter message to represent the fact that all instances belong to a class, and a longer one to represent a spread). 

As previously mentioned, tree classifiers are grown by ``splitting'' nodes, starting at an initial node, the root. Each split at a node corresponds to dividing the instance space with decision boundaries that provide an optimal separation of classes based on some convenient measure of separation such as Information Gain or Gini coefficient, creating corresponding child nodes from which the sub-division process may continue.

The class purity of the distribution at a node is compared with the aggregate class purity of the distributions resulting from a split in order to compute Information Gain. The Information Gains due to several attributes being considered at a node are then compared to find the best one. Simplifying Quinlan's terminology and notation from (p21-22) \parencite{Quinlan:2031749}, the information (or entropy) of a class distribution at a node N is $I_N = \sum_{i=1}^c -p_i \log p_i$, where $p_i$ is the probability observing class $i$ at node $N$ and $c$ is the number of classes. This characterization encapsulates neatly the notion of a ``pure" node having low information---for a one-class distribution, it is easy to observe that information evaluates to $0$.

If node $N$ were now split on attribute $A$ leading to $j$ child nodes, the information contained in each of the child nodes is $I_{N_A^j} = \sum_{i=1}^c -p_i^{N_A^j} \log p_i^{N_A^j}$, and the Information Gain is given by $\sum_j I_{N_A^j} - I_N$. Finding the attribute $A$ that maximises gain is used as a heuristic in C4.5 and similar algorithms to determine the best split.

Classification and Regression Trees (CART) \parencite{breiman1984classification}, which use the Gini impurity instead of Information Gain as a splitting heuristic, \parencite{breiman1984classification} precede C4.5. In this work, we focus our discussion around Information Gain, though it applies equally to Gini-based trees.

Applying C4.5 or CART to a streaming setting is not straightforward; the main problem is that of anytime prediction. How many examples does one need to see before deciding that one has enough data accumulated to create a reliable model? Do we need separate training, validation and test sets in a streaming scenario? How frequently does one need to update this model in order to optimize prediction accuracy? Should one use a sequence of sliding windows? How does one address concept drift in streams? These are the questions that online decision trees address---a set of questions rather different to those centred around data scarcity that apply to the batch paradigm.

\subsection{Online Decision Trees}

In learning from potentially infinite data streams, it is imperative that instances are not stored. One potential solution is to learn in multiple passes with sets of stored instances; this raises the question of what the ideal working instance repository size must be---that is, \textit{how} many instances should be stored at any given time?---a finicky hyperparameter. Such a choice would require a significant space overhead and unduly influence the method's anytime predictions---predictions requested on-demand during the continuous learning process, a standard expectation of online learners.

A one-pass solution, wherein each example is processed exactly once, is ideal for such settings; HoeffdingTree was one of several attempts \parencite{schlimmer1986case, utgoff1989incremental} to provide a one-pass solution, and the first one-pass learner to provide guarantees on deviation of the tree from the batch tree---the hypothetical tree that would be learned if all infinite examples from a stationary distribution were made available at once. Hoeffding Tree uses a statistical test---the Hoeffding Test \parencite{domingos2000mining, hoeffding1963probability}---to determine the most appropriate time to split. Its success may be attributed to the fact that it provided both a one-pass solution and deviation guarantees in the same package.

Work on scalability of batch learners also helped set the foundation for one-pass learning in sequential prediction scenarios. Bootstrapped Optimistic Algorithm for Tree construction (BOAT) \parencite{Gehrke:1999:BDT:304182.304197} represents a typical attempt at learning from a large database that does not use a predictive sequential setting, by sampling fixed size chunks that are used to bootstrap multiple trees. A ``coarse" tree is then extracted, based on the overlapping parts of the bootstrapped trees in terms of split decisions; this tree is further refined to produce a final tree by passing the whole dataset over it. The system is ``incremental" in the sense that it can process additional datasets; and it is responsive to drift in that the system detects when a new dataset requires a change in split criterion at a node through a global assessment of split criterion, and causes a rebuild of the subtree rooted at that node. While key ideas that shape later trees are developed in this work, the sizes of the initial bootstrap samples are arbitrarily chosen, and concurrently the notion of anytime prediction is not entertained---there is no automated way of determining how many examples suffice to build a first reliable tree. Further, the focus is on minimising utilisation of main memory; it is assumed that the database $D$ is available for a corrective step in the algorithm. On the other hand, Hoeffding Tree is truly one-pass, in that it is assumed that an example is seen only once, then discarded. Meanwhile, the RainForest framework \parencite{gehrke2000rainforest} introduces the idea of storing attribute-value-class counts at nodes, which we see in Hoeffding Tree as \textit{node statistics}.

Hoeffding Tree and Hoeffding Adaptive Tree (HAT) are state-of-the-art one-pass incremental learners for stationary and nonstationary streams respectively; we discuss these methods and briefly outline the primary unspecified effects we observe that influence prequential accuracy performance.

\subsubsection{Hoeffding Tree}

Hoeffding Tree \parencite{domingos2000mining} is an online learning strategy that takes as input a stream of instances $(i_1, i_2, ... i_t, i_{t+1}, ...)$ and incrementally builds a decision tree that offers anytime prediction. That is, at any point of time ``t'', a regression or class value prediction can be made based on the examples in the stream up to that point.

The provision of deviation guarantees, and the use of statistical tests (the Hoeffding Test) to make reliable split decisions in a single-pass paradigm made Hoeffding Tree a durable baseline method that won it the KDD Test of Time award in 2015.

Unlike batch decision trees, which process all instances at once, online decision trees need to decide when they are ready to split a node. That is, they need to decide what value of $t$ in the instance stream offers sufficient confidence for a reliable decision boundary to be drawn in input space by splitting a node. Another way of framing this is thus: assuming a stationary distribution, if the entire infinity of instances were filtered down the tree as if in a batch setting, the resulting split is the ideal split---and online decision trees aim to approximate such an ideal split as closely as possible, but because they have to do so in finite time, they need to decide at what time $t$ they have enough confidence that a proposed split for a node matches the ``ideal'' split.

And so: deciding \textit{when} to split a node is the problem that Hoeffding Tree solves in a principled and reliable manner; the Hoeffding Test is used to determine the likelihood of the true best split---as would be obtained by a hypothetical batch tree able to process all infinite examples at once---varying from the best split being considered by the algorithm at a given time. When a certain preset level of confidence is reached at a node, it is split. Terminals do not have children; they are called ``leaf'' nodes. Leaf nodes are evaluated for splits that may result in a better model. 

In Hoeffding Tree, leaf nodes uniquely serve the purpose of ``learning'', in the sense of attempting to find a split point based on examples accumulated at the leaf. Once a learning node has been split on, the ``split node" that replaces it only serves to filter down examples. Leaf nodes that serve as learning nodes update their statistics---and also make predictions. Learning nodes, in Hoeffding Tree, are always leaf nodes.

Statistics that represent the characteristics of the distribution over data attributes and attribute-values are stored in lieu of the data at each node. These node statistics help determine what child nodes will result in a split on a particular attribute at a learning node. They are used to compute test values for the Hoeffding Test; splits are made when a new division of the space by a certain attribute is ascertained to increase the separation power reliably compared to the next best attribute based on the Hoeffding Test applied to Information Gain, Gini coefficient, or other measure of comparative separation due to the attributes.

\begin{algorithm}

	\DontPrintSemicolon
	\SetAlgoLined
	\KwIn{$S$, a sequence of examples,
		\\ \Indp\Indp $\mathbf{X}$, a set of discrete attributes,
		\\ G(.), a split evaluation function
		\\ $\delta$, one minus the desired probability of choosing the correct attribute at any given node
	}
	\KwOut{$HT$, a decision tree.}
	\Begin{
		Let HT be a tree with a single leaf $l_1$ (the $root$).\;
		Let $\mathbf{X_1} = \mathbf{X} \cup X_{\emptyset}$.\;
		Let $\overline{G_1}(X_{\emptyset})$ be the $\overline{G}$ obtained by predicting the most
		frequent class in $S$\;
		\ForEach {class $y_k$}{
			\ForEach { value $x_{ij}$ of each attribute $X_i \in \mathbf{X}$}{
				Let $n_{ijk}(l_1) = 0$\;
			}
		}
		
		\ForEach {example $(\vec{x},y)$ in S}{
			Sort $(\vec{x},y)$ into a leaf $l$ using $HT$\;
			\ForEach {$x_{ij}$ in $\vec{x}$ such that $X_i \in \mathbf{X}_l$}{
				Increment $n_{ijk}(l)$\;
			}
			Label $l$ with the majority class among the examples seen so far at $l$\;
			\If {the examples seen so far at $l$ are not all of the same class}{
				Compute $\overline{G_l}(X_i)$ for each attribute $X_i \in \mathbf{X}_l - \{X_{\emptyset}\}$ using the counts $n_{ijk}(l)$\;
				Let $X_a$ be the attribute with highest  $\overline{G_l}$\;
				Let $X_b$ be the attribute with second-highest  $\overline{G_l}$\;
				Compute $\epsilon$ using Equation \ref{eq:1}\;
				\If {$\overline{G_l}(X_a) - \overline{G_l}(X_b) > \epsilon$ and $X_a \neq X_{\emptyset}$}{
					Replace $l$ by an internal node that splits on $X_a$
					\ForEach {branch of the split}{
						Add a new leaf $l_m$ and let $\mathbf{X}_m = \mathbf{X} - \{X_\emptyset\}$
						Let $\overline{G}_m(X_\emptyset)$ be the $\overline{G}$ obtained by predicting the most frequent class at $l_m$
						\ForEach {class $y_k$ and each value $X_{ij}$ of each attribute $X_i \in \mathbf{X}_m - \{X_\emptyset\}]\}$}{
							Let $n_{ijk}(l_m) = 0$	
						}
					}
				}
			}			
		}
		Return HT
	}

	\caption{Hoeffding Tree, Domingos \& Hulten (2000)
		\Indp\Indp\Indp\Indp  \small{--Reproduced verbatim from original-- }
	} 
	\label{table:vfdt}
\end{algorithm}

\subsubsection{Hoeffding Adaptive Tree}

Hoeffding Adaptive Tree (HAT) is an adaptive online decision tree based on Hoeffding Tree. HAT starts growing an alternate subtree when concept drift is detected at a node by ADWIN \parencite{bifet2007learning} or another change detector, and replaces the original subtree with an alternate when the error from the alternate is lower than the error from the main subtree. Till date, it remains the best performing single online tree on our expanded concept drift testbench in terms of prequential error; we experimented with several strategies to improve performance, such as allowing alternates to vote as a form of lookahead and weighting examples at subtrees as a form of subtree-level example boosting, only to find that unspecified features already accounted for them to various degrees. In Section \ref{hat}, we explain what these unspecified features are and how they result in superior performance, isolating strategies that boost prequential accuracy.

\section{Unspecified and Emergent Behaviors in the MOA implementation of Hoeffding Tree}
\label{ht}

An online strategy designed for concept drift adapts to the stream in order to predict as well as possible as concepts change; an online strategy designed for learning from stationary streams assumes an overall unchanging distribution, and prioritizes minimizing deviation from  an ideal batch learner that has access to the entire stream of infinite size at once. A learner designed for a stationary stream can be expected to be stable on the stability-plasticity \parencite{grossberg1988nonlinear, hoens2012learning} spectrum, as it aims to mitigate a reactive change in the model on exposure to noise so it can capture the unchanging concept.

HoeffdingTree---the state-of-the-art online learning tree that has served as a base procedure for the development of methods for learning from online streams with concept drift---is surprisingly responsive to abrupt drifts, in spite of having been designed for learning from stationary streams and for very high stability. Abrupt drifts are large shifts in concept over a relatively short period of time \parencite{Gama:eval_survey, webb2016characterizing}---in our experimental setup, abrupt drifts are considered instantaneous, with the data-generating distribution (``concept") switching instantly to a new one.

Closer inspection reveals several factors promoting a form of response to drift that we call \textit{amnesia}. By \textit{amnesia} we refer to the ability of a learner to update its model to better capture the present state of the stream by diminishing the influence of earlier examples---effectively, forgetting them to some degree.

One of the factors leading to the excellent drift response of Hoeffding Tree in any reasonable implementation is inadvertent inbuilt amnesia---as instances are not stored, and there is no effective way to exactly determine node statistics in child nodes resulting from a split, node statistics are simply initialised to zero for convenience, effectively deleting history upon splitting. We elaborate on this in Section \ref{ht-amnesia}.

A second factor that is unique to the MOA interpretation of Hoeffding Tree is that it was possible to re-split on a used nominal attribute, again contributing to amnesia when it was most useful. We explain this in greater depth in Section \ref{ht-resplitting}.

A further simplification, which we explain in Section \ref{ht-infogain}, in the MOA implementation---one which contributes to time efficiency without changing time complexity or producing a discernible change in prequential accuracy---is that Information Gains are not averaged; periodic infogain computations are considered good enough approximations---as they indeed turn out to be in practice. This is a significant optimization that could be written into other HoeffdingTree implementations.

Put together, these factors lead to a significance performance gain over a base Very Fast Decision Tree (VFDT), the implementation of the theoretical Hoeffding Tree construct. We use the MOA implementation, though some significant factors are also part of the original VFML implementation. (Section \ref{ht-all}).

\subsection{Experimental Setup}
\label{exp_setup}

We use the Massive Online Analysis (MOA) \parencite{bifet2010moa} framework for our experimentation. MOA provides implementations of Hoeffding Tree and Hoeffding Adaptive Tree, and a number of concept drift streams. For more detailed analysis, we add streams generated by a variant of the stream generator from \parencite{webb2016characterizing}, which gives us fine control over the generation of abrupt drift. We use predictive sequential accuracy---prequential accuracy---as our performance measure. This is the cumulative accuracy from a setting where a prediction is offered for each instance by the learner, and the true value is made available following the prediction. We focus on the classification problem.

As a base for our experimentation we use implementations of VFDT and HAT stripped of the unspecified features we have found in the MOA and VFML implementations. All references to ``VFDT'' or ``HAT'' in our charts and tables refer to an implementation of the core Hoeffding Tree (Algorithm \ref{table:vfdt}) as described in \parencite{domingos2000mining} and the Hoeffding Adpaptive Tree algorithm \parencite{bifet2009adaptive} without any of the unspecified features that have entered the implementations of VFDT and HAT in the form of engineering artifacts. 

We have added as options to these base implementions each of the unspecified features we have identified in the MOA implementations and in the following sections systematically investigate their impact on prequential accuracy.

\subsection{Inherent Amnesia}
\label{ht-amnesia}

HoeffdingTree, as a stream learner, is designed to operate with a low, finite memory requirement. Examples are not stored, but statistics about examples are at each node, ``node statistics'' that help determine the resultant class distributions when a chosen attribute is split upon. Node statistics are simply representations of the data collected at the node---frequency counts for nominal attribute values, and appropriate statistics such as mean for real attribute values. This allows determination of the best split attribute---one can choose the attribute that results in maximal class separation.

A critical side effect of the usage of node statistics is that freshly created nodes have to start off with \textit{no} node statistics, as there does not appear to be an obviously sound way of redistributing node statistics from the parent without storing examples---one needs the examples in order to be able to determine an appropriate redistribution of node statistics (by filtering examples down the tree). 

Though Hoeffding Tree was designed for stationary streams, the consequence of new nodes being created without statistics turns out to be a significant unplanned advantage in the scenario with concept drift; a tree algorithm may ``recover" from a concept drift (that is, update its model to be current) simply by splitting, leading to amnesia that erases the previous concept---thus modeling the novel concept better than expected after a concept drift.

For example: in an extreme case, suppose previously that all 1000 examples that reached a leaf belonged to class A. After drift, all examples reaching the leaf belong to class B. Simply by splitting on a now irrelevant attribute, the tree can forget the 1000 examples of class A in the node statistics. A further split may be justified based on the new separable node statistics, and this would render a child with a class distribution derived from these fresh node statistics that classify concept B well.

In order to demonstrate this effect, we built a version of VFDT that does not forget instances, so that the node statistics $n_{ijk}$ in the children are initialised to counts from all the examples $i$ in the stream to date that filter into the freshly created child nodes, instead of being set to zero. 

We then set up a simple noise-free test stream with drift in the conditional distribution P(Y$|$X) with 5 attributes, 5 nominal values per attribute and 5 classes. The stream is from \parencite{webb2016characterizing} and is initialised thus: first, a ``starting" distribution is created. A random distribution over the covariates is created by drawing attribute-value probabilities from a Gamma(1,1) function and normalizing them per attribute (every attribute gets a value in every instance); and every possible combination of values of the covariates is assigned a random class by drawing an index from a uniform distribution. Next, a second, ``final" distribution after drift is created by randomly changing the class assignment, ensuring we pick a class that is not the one that is already chosen. The proportion of classes to be changed is provided as an input for drift magnitude.

\begin{figure}[t]
	\includegraphics[width=12cm]{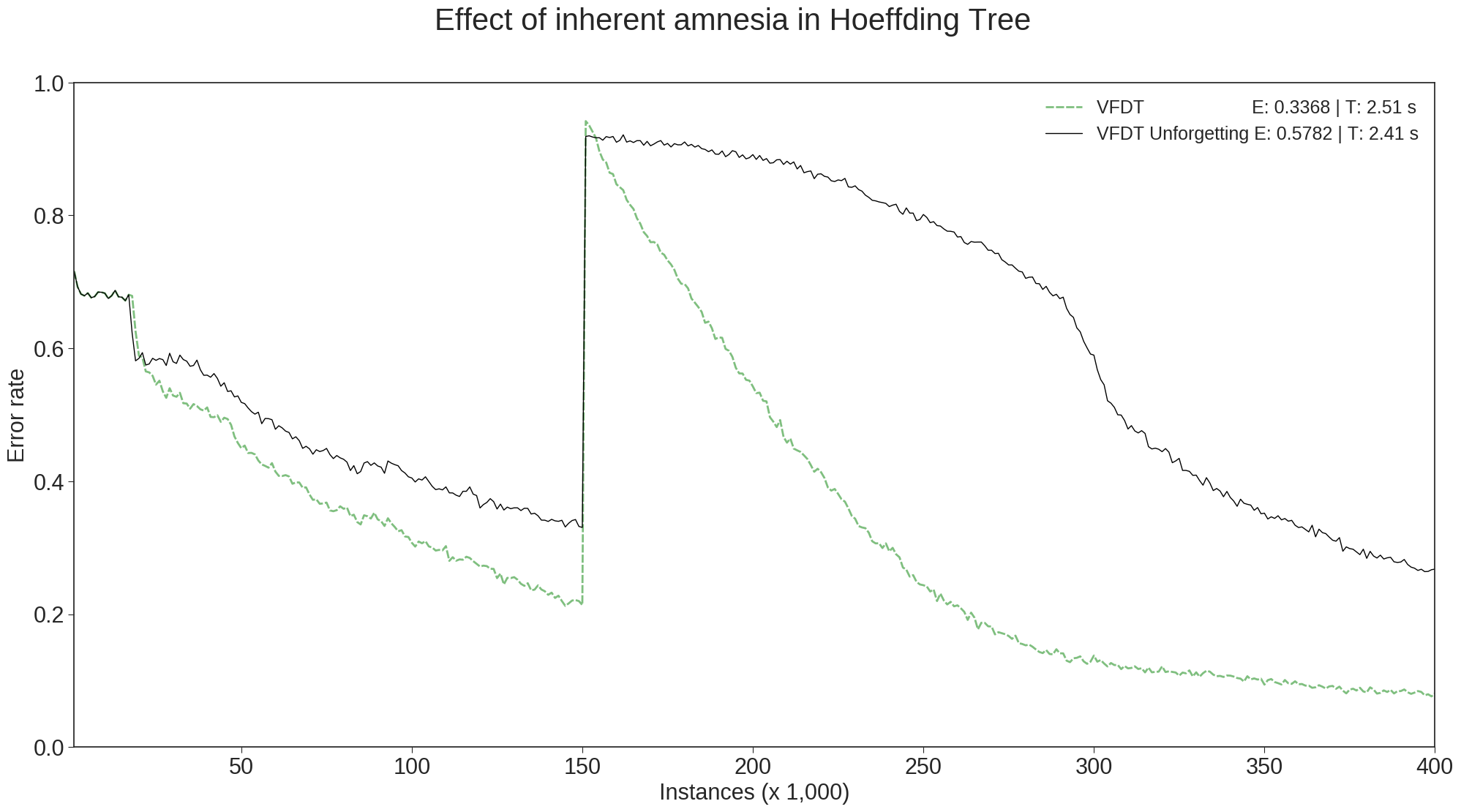}
	\centering
	\caption{Plot showing a significant improvement in drift response resulting from amnesia inherent to HoeffdingTree. A single abrupt drift occurs at $t=150,000$. VFDT shows a far better recovery in prequential error following the drift. `E' and `T' are error and time, respectively, averaged every 1000 steps over 10 randomised streams.}
	\label{fig:inherentAmnesia}
\end{figure}

Figure \ref{fig:inherentAmnesia} plots the prequential accuracies of unmodified Hoeffding Tree and an ``eidetic'' Hoeffding Tree that retains instances for initialization of the node statistics $n_{ijk}$ on a synthetic stream with 5 classes, 5 nominal attributes, and 5 values per attribute. An abrupt drift occurs at $t=150,000$ with the highest possible magnitude of 1.0. The figure shows that due to the amnesia inherent in HoeffdingTree, response to drift in terms of recovery in error rate following the abrupt drift is far quicker than it is without inherent amnesia.

\subsection{MOA implementation: Resplitting on attributes}
\label{ht-resplitting}

If a nominal attribute that has already been used to split on a decision path is reused to split, we should achieve no advantage at all and only the disadvantage of making the tree larger. Because the value for the previously used nominal attribute has already been decided the first time it was split upon, ``splitting'' on it again will make no difference whatsoever to the sequence of decisions down the path, and it will in no way discriminate any set of instances accumulated at the node---all instances will contain the same value of the reused attribute! As a result,  all examples will pass down the same branch below the split---hence the distribution at the new leaf should be identical to that at the current one. 

Thus, when a leaf node evaluates multiple split options, it should normally be the case that a previously used nominal attribute will offer zero information gain---it has already been conditioned upon and hence will all have the same value. 

However, VFDT was designed for stationary streams; at least in the MOA implementation, the introduction of concept drift leads to computations of Information Gain taking negative values. Because the correctly calculated Information Gain values of $0$ obtained through reusing an attribute are greater than such negative values, previously used attributes are enabled for reuse---that is, the implementation interpretation allows re-``splitting" on the same attribute---with the same previously used attribute-value being assigned (as the attribute has already taken it's value, and with any other value an example would end up elsewhere in the tree).

This re-``splitting" behavior has not been found to manifest in drift-free streams. Concept drift triggers it, and the end result is very interesting---amnesia. In the freshly created leaf node on a resplit on a previously used attribute, node statistics are zeroed as a consequence of the effect described in Section \ref{ht-amnesia}; this amnesia helps concept drift adaptation, again, by allowing the tree to model a newer concept at any future leaves without the weight of the previous concept as the class distribution used to predict is derived from the parent's node statistics, which have been cleared. The interaction between inherent amnesia and concept-drift triggered ``resplitting'' leads over all to a better drift response as can be seen in Table \ref{table101}, which shows an overwhelming win rate on our testbench for VFDT that allows resplitting over our base VFDT that does not. The number of wins with the resplitting strategy over without it is significant in itself, but further consequence is found in how this interacts with other unspecified features in order to produce a universal ``win", as demonstrated in Section \ref{ht-all}.

As ``re-splitting'' proves to be a powerful strategy for responding to concept drift, it might be instructive to experiment with variations of initializing/zeroing node statistics when a drift is detected or suspected in adaptive methods designed specifically for concept drift.  We provide an example of doing so in Section \ref{ht-evisceration}.

\begin{table}[!ht]
	
	\caption{\label{table101}Performance of the ``resplitting" strategy: A previously used attribute is ``reused", effectively creating a single child with a clean node statistics register}
	\scriptsize
	\centering
	\makebox[0.9\textwidth]{
		
		\begin{tabular}{m{12cm}?m{1.4cm}?m{1.4cm}}
			\toprule
			\multicolumn{1}{c?}{\textbf{Streams}} &  \multicolumn{1}{m{1.4cm}}{\textbf{VFDT}} &
			\textbf{VFDT with resplitting strategy}\\
			\midrule
			RecurrentConceptDriftStream -x 200000 -y 200000 -z 100 -s (AgrawalGenerator -f 2 -i 2) -d (AgrawalGenerator -f 3 -i 3) & 0.20846 & \textbf{0.19894} \\
RecurrentConceptDriftStream -x 200000 -y 200000 -z 100 -s (RandomTreeGenerator -r 1 -i 1) -d (RandomTreeGenerator -r 2 -i 2) & 0.22404 & \textbf{0.22151} \\
RecurrentConceptDriftStream -x 200000 -y 200000 -z 100 -s (SEAGenerator -f 2 -i 2) -d (SEAGenerator -f 3 -i 3) & 0.15251 & \textbf{0.15239} \\
RecurrentConceptDriftStream -x 200000 -y 200000 -z 100 -s (STAGGERGenerator -i 2 -f 2) -d (STAGGERGenerator -i 3 -f 3) & 0.1882 & \textbf{0.00699} \\
HyperplaneGenerator -k 10 -t 0.0001 -i 2 & 0.11566 & \textbf{0.11455} \\
HyperplaneGenerator -k 10 -t 0.001 -i 2 & 0.16785 & \textbf{0.16458} \\
HyperplaneGenerator -k 10 -t 0.01 -i 2 & 0.1713 & \textbf{0.1685} \\
HyperplaneGenerator -k 5 -t 0.0001 -i 2 & 0.1074 & \textbf{0.10714} \\
HyperplaneGenerator -k 5 -t 0.001 -i 2 & 0.16309 & \textbf{0.16005} \\
HyperplaneGenerator -k 5 -t 0.01 -i 2 & 0.17108 & \textbf{0.16579} \\
LEDGeneratorDrift -d 1 -i 2 & \textit{\textbf{0.26093}} & \textit{\textbf{0.26093}} \\
LEDGeneratorDrift -d 3 -i 2 & \textit{\textbf{0.26093}} & \textit{\textbf{0.26093}} \\
LEDGeneratorDrift -d 5 -i 2 & \textit{\textbf{0.26093}} & \textit{\textbf{0.26093}} \\
LEDGeneratorDrift -d 7 -i 2 & \textit{\textbf{0.26093}} & \textit{\textbf{0.26093}} \\
RandomRBFGeneratorDrift -s 0.0001 -k 10 -i 2 -r 2 & 0.11462 & \textbf{0.11381} \\
RandomRBFGeneratorDrift -s 0.0001 -k 50 -i 2 -r 2 & 0.2858 & \textbf{0.28486} \\
RandomRBFGeneratorDrift -s 0.001 -k 10 -i 2 -r 2 & 0.13821 & \textbf{0.1377} \\
RandomRBFGeneratorDrift -s 0.001 -k 50 -i 2 -r 2 & 0.40874 & \textbf{0.40583} \\
WaveformGeneratorDrift -d 1 -i 2 -n & \textit{\textbf{0.16284}} & \textit{\textbf{0.16284}} \\
WaveformGeneratorDrift -d 3 -i 2 -n & \textit{\textbf{0.16284}} & \textit{\textbf{0.16284}} \\
WaveformGeneratorDrift -d 5 -i 2 -n & \textit{\textbf{0.16284}} & \textit{\textbf{0.16284}} \\
WaveformGeneratorDrift -d 7 -i 2 -n & \textit{\textbf{0.16284}} & \textit{\textbf{0.16284}} \\
AbruptDriftGenerator -c  -o 1.0 -z 2 -n 2 -v 2 -r 2 -b 200000 -d Recurrent & 0.35403 & \textbf{0.00261} \\
AbruptDriftGenerator -c  -o 1.0 -z 3 -n 2 -v 2 -r 2 -b 200000 -d Recurrent & 0.37862 & \textbf{0.00325} \\
AbruptDriftGenerator -c  -o 1.0 -z 3 -n 3 -v 2 -r 2 -b 200000 -d Recurrent & 0.3504 & \textbf{0.01316} \\
AbruptDriftGenerator -c  -o 1.0 -z 3 -n 3 -v 3 -r 2 -b 200000 -d Recurrent & 0.36505 & \textbf{0.0582} \\
AbruptDriftGenerator -c  -o 1.0 -z 3 -n 3 -v 4 -r 2 -b 200000 -d Recurrent & 0.39687 & \textbf{0.15073} \\
AbruptDriftGenerator -c  -o 1.0 -z 3 -n 3 -v 5 -r 2 -b 200000 -d Recurrent & 0.39622 & \textbf{0.23007} \\
AbruptDriftGenerator -c  -o 1.0 -z 4 -n 2 -v 2 -r 2 -b 200000 -d Recurrent & 0.33416 & \textbf{0.00959} \\
AbruptDriftGenerator -c  -o 1.0 -z 4 -n 4 -v 4 -r 2 -b 200000 -d Recurrent & 0.40671 & \textbf{0.34738} \\
AbruptDriftGenerator -c  -o 1.0 -z 5 -n 2 -v 2 -r 2 -b 200000 -d Recurrent & 0.3309 & \textbf{0.00937} \\
AbruptDriftGenerator -c  -o 1.0 -z 5 -n 5 -v 5 -r 2 -b 200000 -d Recurrent & 0.46461 & \textbf{0.46352} \\
\bottomrule &  &  \\
\begin{tabularx}{\linewidth}{Xr}
		A \textbf{bold} value indicates higher accuracy, and \textit{\textbf{bold italics}} indicate a tie.  & \textbf{Unique Wins}
		\end{tabularx} & \textbf{0} & \textbf{24} \\
\cmidrule[0.4pt](lr){2-3} &  &  \\
\begin{tabularx}{\linewidth}{Xr} The test is a one-tailed binomial test to determine the probability that the strategy in the  
             rightmost column would achieve so many wins if wins and losses were equiprobable. & \textbf{Test Statistics} \end{tabularx} & \textbf{p-value: is $<$ 0.00001} & \textbf{Confidence Interval:  0.88265 --- 1} \\

			\bottomrule
		\end{tabular}
	}
\end{table}

\subsection{Infogain approximations}
\label{ht-infogain}

Hoeffding tree is premised upon offering rationalised splits in streaming scenarios using the Hoeffding Inequality as a test to determine whether a potential split is likely to be the long-term choice in a theoretical batch tree built on an infinite stationary stream. Hoeffding test is applied with the computed prospective infogain differences between the top two attributes due to a potential split at timestep $t$, $\Delta G_t$, as the random variables.

The standard use of Hoeffding's inequality would involve taking the mean of all computed infogain differences  $\Delta G_t$ over all timesteps \parencite{domingos2000mining, hoeffding1963probability}. That is, every time the infogain difference $\Delta G_t$ is computed, it is fed into the mean value $\overline{\Delta G}_t$ comprising all $\Delta G_t$ up until that timestep---the value used in the Hoeffding Test to determine the relative superiority of an attribute. (We use timestep subscripts to clarify the meanings of the random variables. Random variables are often assumed to be ``taking values in sequence"; this is a misconception. A random variable is machinery to represent uncertainty about a particular event, and multiple events correspond to different random variables, even if drawn from the same distribution---the random variable does not take ``one value after another").

In practice, for the sake of efficiency, infogain is only computed at set intervals, such as once every 200 timesteps. This implies, for example, that over 10 invocations of the $\overline{\Delta G}_t$ computations over $2,000$ timesteps, one would only have $n=10$ of the random variables $\overline{\Delta G}_t$ to average. Having fewer $\overline{\Delta G}_t$ to average, would mean a potentially longer wait for VFDT for the Hoeffding Test to become significant.

Both the MOA implementation \parencite{bifet2010moa} and the original Very Fast Machine Learning (VFML) implementation \parencite{domingos2000mining} take $n$ to be the number of examples (not the number of $\overline{\Delta G}_t$ computations), and use the most recently computed $\Delta G_t$ divided by $n$ as a proxy for the average $\overline{\Delta G}_t$, at variance with the published algorithm in \parencite{domingos2000mining}.

In the stationary case, this is perfectly reasonable to do. After all, we do expect $\overline{\Delta G}_t$ to converge in the long run, and we do not expect unusually great fluctuations after having observed each instance. Therefore, we benefit from quicker decisions in building tree structure as $n$ grows large quickly in step with the number of examples and not in step with infogain computations.


In the drifting scenario, the larger $n$ implies that the much more variable instantaneous infogain computations would be accepted faster than theoretically expected by the algorithm in situations where drift induces the gap between the best candidate split attribute and the second best one to widen. In contrast, averaging would slow down response as one uses a ``fuzzy" average of potentially unhelpful gains over a small $n$. Table \ref{table102} shows us how the unspecified strategy of infogain approximation fares in terms of increasing prequential accuracy---with 19 wins to 10 losses when approximation is not used. Thus there is some evidence to support infogain approximation having a positive effect, but it is not conclusive in the case of VFDT.

\begin{table}[!ht]
	
	\caption{\label{table102}Performance of the infogain approximation strategy}
	\scriptsize
	\centering
	\makebox[0.6\textwidth]{
		
		\begin{tabular}{m{12cm}?m{1.4cm}?m{1.4cm}}
			\toprule
			\multicolumn{1}{c?}{\textbf{Streams}} &  \multicolumn{1}{m{1.4cm}}{\textbf{VFDT without infogain approximation}} &
			\textbf{VFDT with infogain approximation}\\
			\midrule
			RecurrentConceptDriftStream -x 200000 -y 200000 -z 100 -s (AgrawalGenerator -f 2 -i 2) -d (AgrawalGenerator -f 3 -i 3) & 0.20846 & \textbf{0.20774} \\
RecurrentConceptDriftStream -x 200000 -y 200000 -z 100 -s (RandomTreeGenerator -r 1 -i 1) -d (RandomTreeGenerator -r 2 -i 2) & 0.22404 & \textbf{0.22367} \\
RecurrentConceptDriftStream -x 200000 -y 200000 -z 100 -s (SEAGenerator -f 2 -i 2) -d (SEAGenerator -f 3 -i 3) & 0.15251 & \textbf{0.15228} \\
RecurrentConceptDriftStream -x 200000 -y 200000 -z 100 -s (STAGGERGenerator -i 2 -f 2) -d (STAGGERGenerator -i 3 -f 3) & \textit{\textbf{0.1882}} & \textit{\textbf{0.1882}} \\
HyperplaneGenerator -k 10 -t 0.0001 -i 2 & \textbf{0.11566} & 0.11576 \\
HyperplaneGenerator -k 10 -t 0.001 -i 2 & \textbf{0.16785} & 0.1681 \\
HyperplaneGenerator -k 10 -t 0.01 -i 2 & \textbf{0.1713} & 0.17202 \\
HyperplaneGenerator -k 5 -t 0.0001 -i 2 & \textbf{0.1074} & 0.10745 \\
HyperplaneGenerator -k 5 -t 0.001 -i 2 & 0.16309 & \textbf{0.16251} \\
HyperplaneGenerator -k 5 -t 0.01 -i 2 & \textbf{0.17108} & 0.17114 \\
LEDGeneratorDrift -d 1 -i 2 & 0.26093 & \textbf{0.26086} \\
LEDGeneratorDrift -d 3 -i 2 & 0.26093 & \textbf{0.26086} \\
LEDGeneratorDrift -d 5 -i 2 & 0.26093 & \textbf{0.26086} \\
LEDGeneratorDrift -d 7 -i 2 & 0.26093 & \textbf{0.26086} \\
RandomRBFGeneratorDrift -s 0.0001 -k 10 -i 2 -r 2 & 0.11462 & \textbf{0.11446} \\
RandomRBFGeneratorDrift -s 0.0001 -k 50 -i 2 -r 2 & 0.2858 & \textbf{0.28565} \\
RandomRBFGeneratorDrift -s 0.001 -k 10 -i 2 -r 2 & \textbf{0.13821} & 0.13867 \\
RandomRBFGeneratorDrift -s 0.001 -k 50 -i 2 -r 2 & \textbf{0.40874} & 0.40887 \\
WaveformGeneratorDrift -d 1 -i 2 -n & 0.16284 & \textbf{0.16276} \\
WaveformGeneratorDrift -d 3 -i 2 -n & 0.16284 & \textbf{0.16276} \\
WaveformGeneratorDrift -d 5 -i 2 -n & 0.16284 & \textbf{0.16276} \\
WaveformGeneratorDrift -d 7 -i 2 -n & 0.16284 & \textbf{0.16276} \\
AbruptDriftGenerator -c  -o 1.0 -z 2 -n 2 -v 2 -r 2 -b 200000 -d Recurrent & 0.35403 & \textbf{0.35402} \\
AbruptDriftGenerator -c  -o 1.0 -z 3 -n 2 -v 2 -r 2 -b 200000 -d Recurrent & \textit{\textbf{0.37862}} & \textit{\textbf{0.37862}} \\
AbruptDriftGenerator -c  -o 1.0 -z 3 -n 3 -v 2 -r 2 -b 200000 -d Recurrent & \textbf{0.3504} & 0.35104 \\
AbruptDriftGenerator -c  -o 1.0 -z 3 -n 3 -v 3 -r 2 -b 200000 -d Recurrent & 0.36505 & \textbf{0.36503} \\
AbruptDriftGenerator -c  -o 1.0 -z 3 -n 3 -v 4 -r 2 -b 200000 -d Recurrent & 0.39687 & \textbf{0.39678} \\
AbruptDriftGenerator -c  -o 1.0 -z 3 -n 3 -v 5 -r 2 -b 200000 -d Recurrent & \textbf{0.39622} & 0.39635 \\
AbruptDriftGenerator -c  -o 1.0 -z 4 -n 2 -v 2 -r 2 -b 200000 -d Recurrent & \textit{\textbf{0.33416}} & \textit{\textbf{0.33416}} \\
AbruptDriftGenerator -c  -o 1.0 -z 4 -n 4 -v 4 -r 2 -b 200000 -d Recurrent & 0.40671 & \textbf{0.40632} \\
AbruptDriftGenerator -c  -o 1.0 -z 5 -n 2 -v 2 -r 2 -b 200000 -d Recurrent & \textbf{0.3309} & 0.3312 \\
AbruptDriftGenerator -c  -o 1.0 -z 5 -n 5 -v 5 -r 2 -b 200000 -d Recurrent & 0.46461 & \textbf{0.46106} \\
\bottomrule &  &  \\
\begin{tabularx}{\linewidth}{Xr}
		A \textbf{bold} value indicates higher accuracy, and \textit{\textbf{bold italics}} indicate a tie.  & \textbf{Unique Wins}
		\end{tabularx} & \textbf{10} & \textbf{19} \\
\cmidrule[0.4pt](lr){2-3} &  &  \\
\begin{tabularx}{\linewidth}{Xr} The test is a one-tailed binomial test to determine the probability that the strategy in the  
             rightmost column would achieve so many wins if wins and losses were equiprobable. & \textbf{Test Statistics} \end{tabularx} & \textbf{p-value: 0.06802} & \textbf{Confidence Interval:  0.48573 --- 1} \\

			\bottomrule
		\end{tabular}
	}
\end{table}

\subsection{Using number of timesteps instead of weight seen at leaves to determine split evaluation}
\label{ht-nodetime}

Each instance is assumed to be labeled with a class label $i$, and a separate count $c_i$ is maintained for each class label at each node, leading to a class distribution.

When a split occurs on a node, child nodes are initialised with a class distribution $Y$ derived from that of the parent node---the parent class counts $c_i$ are distributed among the child nodes at their inception. Thus child leaf nodes are initialised with a class distribution that is useful for making predictions from the point they are created.

Because of the class weight due to initialisation, the class weight $\sum c_i$ for classes $i$ seen at leaves is generally larger than the actual number of instances seen at the leaf. Thus if the summed class weight $\sum c_i$ is used as a proxy for the actual number of instances $N$ seen at the leaf, we will generally end up using a larger value than the actual number of instances seen at the leaf. In other words, the aggregate of the counts of the classes should add up to the number of instances observed (assuming every instance is labeled)---\textit{plus} the counts carried over from the parents at the time of the split.

Now, the application of the Hoeffding Test at a node is based on the number of examples seen at a node. Because $\sum c_i$ is generally larger than $N$, using $\sum c_i$ for the Hoeffding Test in lieu of $N$ helps increase test confidence faster by sending the signal that a larger number of examples than actually observed have been observed. Because the newly created leaf already has node statistics $n_{ijk}$ set to zero, it should respond more easily to change by splitting. Weighted examples may also arise from streams or learners.

Table \ref{table103} shows us how this affects drift streams. The effect is minor and we do not find it particularly significant.

\begin{table}[!ht]
	
	\caption{\label{table103}Using weightSeen instead of number of timesteps}
	\scriptsize
	\centering
	\makebox[0.6\textwidth]{
		
		\begin{tabular}{m{12cm}?m{1.4cm}?m{1.4cm}}
			\toprule
			\multicolumn{1}{c?}{\textbf{Streams}} &  \multicolumn{1}{m{1.4cm}}{\textbf{VFDT with number of timesteps}} &
			\textbf{VFDT with weightSeen}\\
			\midrule
			RecurrentConceptDriftStream -x 200000 -y 200000 -z 100 -s (AgrawalGenerator -f 2 -i 2) -d (AgrawalGenerator -f 3 -i 3) & 0.20846 & \textbf{0.2078} \\
RecurrentConceptDriftStream -x 200000 -y 200000 -z 100 -s (RandomTreeGenerator -r 1 -i 1) -d (RandomTreeGenerator -r 2 -i 2) & 0.22404 & \textbf{0.22403} \\
RecurrentConceptDriftStream -x 200000 -y 200000 -z 100 -s (SEAGenerator -f 2 -i 2) -d (SEAGenerator -f 3 -i 3) & \textbf{0.15251} & 0.15253 \\
RecurrentConceptDriftStream -x 200000 -y 200000 -z 100 -s (STAGGERGenerator -i 2 -f 2) -d (STAGGERGenerator -i 3 -f 3) & \textit{\textbf{0.1882}} & \textit{\textbf{0.1882}} \\
HyperplaneGenerator -k 10 -t 0.0001 -i 2 & 0.11566 & \textbf{0.11565} \\
HyperplaneGenerator -k 10 -t 0.001 -i 2 & \textbf{0.16785} & 0.16801 \\
HyperplaneGenerator -k 10 -t 0.01 -i 2 & \textbf{0.1713} & 0.17145 \\
HyperplaneGenerator -k 5 -t 0.0001 -i 2 & \textbf{0.1074} & 0.10744 \\
HyperplaneGenerator -k 5 -t 0.001 -i 2 & 0.16309 & \textbf{0.16287} \\
HyperplaneGenerator -k 5 -t 0.01 -i 2 & 0.17108 & \textbf{0.17075} \\
LEDGeneratorDrift -d 1 -i 2 & \textit{\textbf{0.26093}} & \textit{\textbf{0.26093}} \\
LEDGeneratorDrift -d 3 -i 2 & \textit{\textbf{0.26093}} & \textit{\textbf{0.26093}} \\
LEDGeneratorDrift -d 5 -i 2 & \textit{\textbf{0.26093}} & \textit{\textbf{0.26093}} \\
LEDGeneratorDrift -d 7 -i 2 & \textit{\textbf{0.26093}} & \textit{\textbf{0.26093}} \\
RandomRBFGeneratorDrift -s 0.0001 -k 10 -i 2 -r 2 & 0.11462 & \textbf{0.11458} \\
RandomRBFGeneratorDrift -s 0.0001 -k 50 -i 2 -r 2 & \textbf{0.2858} & 0.2862 \\
RandomRBFGeneratorDrift -s 0.001 -k 10 -i 2 -r 2 & \textbf{0.13821} & 0.13852 \\
RandomRBFGeneratorDrift -s 0.001 -k 50 -i 2 -r 2 & 0.40874 & \textbf{0.40853} \\
WaveformGeneratorDrift -d 1 -i 2 -n & 0.16284 & \textbf{0.16274} \\
WaveformGeneratorDrift -d 3 -i 2 -n & 0.16284 & \textbf{0.16274} \\
WaveformGeneratorDrift -d 5 -i 2 -n & 0.16284 & \textbf{0.16274} \\
WaveformGeneratorDrift -d 7 -i 2 -n & 0.16284 & \textbf{0.16274} \\
AbruptDriftGenerator -c  -o 1.0 -z 2 -n 2 -v 2 -r 2 -b 200000 -d Recurrent & \textit{\textbf{0.35403}} & \textit{\textbf{0.35403}} \\
AbruptDriftGenerator -c  -o 1.0 -z 3 -n 2 -v 2 -r 2 -b 200000 -d Recurrent & \textit{\textbf{0.37862}} & \textit{\textbf{0.37862}} \\
AbruptDriftGenerator -c  -o 1.0 -z 3 -n 3 -v 2 -r 2 -b 200000 -d Recurrent & \textit{\textbf{0.3504}} & \textit{\textbf{0.3504}} \\
AbruptDriftGenerator -c  -o 1.0 -z 3 -n 3 -v 3 -r 2 -b 200000 -d Recurrent & \textit{\textbf{0.36505}} & \textit{\textbf{0.36505}} \\
AbruptDriftGenerator -c  -o 1.0 -z 3 -n 3 -v 4 -r 2 -b 200000 -d Recurrent & \textit{\textbf{0.39687}} & \textit{\textbf{0.39687}} \\
AbruptDriftGenerator -c  -o 1.0 -z 3 -n 3 -v 5 -r 2 -b 200000 -d Recurrent & \textit{\textbf{0.39622}} & \textit{\textbf{0.39622}} \\
AbruptDriftGenerator -c  -o 1.0 -z 4 -n 2 -v 2 -r 2 -b 200000 -d Recurrent & \textit{\textbf{0.33416}} & \textit{\textbf{0.33416}} \\
AbruptDriftGenerator -c  -o 1.0 -z 4 -n 4 -v 4 -r 2 -b 200000 -d Recurrent & \textit{\textbf{0.40671}} & \textit{\textbf{0.40671}} \\
AbruptDriftGenerator -c  -o 1.0 -z 5 -n 2 -v 2 -r 2 -b 200000 -d Recurrent & \textit{\textbf{0.3309}} & \textit{\textbf{0.3309}} \\
AbruptDriftGenerator -c  -o 1.0 -z 5 -n 5 -v 5 -r 2 -b 200000 -d Recurrent & \textit{\textbf{0.46461}} & \textit{\textbf{0.46461}} \\
\bottomrule &  &  \\
\begin{tabularx}{\linewidth}{Xr}
		A \textbf{bold} value indicates higher accuracy, and \textit{\textbf{bold italics}} indicate a tie.  & \textbf{Unique Wins}
		\end{tabularx} & \textbf{6} & \textbf{11} \\
\cmidrule[0.4pt](lr){2-3} &  &  \\
\begin{tabularx}{\linewidth}{Xr} The test is a one-tailed binomial test to determine the probability that the strategy in the  
             rightmost column would achieve so many wins if wins and losses were equiprobable. & \textbf{Test Statistics} \end{tabularx} & \textbf{p-value: 0.16615} & \textbf{Confidence Interval:  0.41971 --- 1} \\

			\bottomrule
		\end{tabular}
	}
\end{table}

\subsection{Putting it all together}
\label{ht-all}

Individually, the unspecified strategies from Sections \ref{ht-amnesia},\ref{ht-resplitting}, \ref{ht-infogain}, \ref{ht-nodetime} sometimes lead to significant performance increases; put together, this gap widens so that a universal ``win" is obtained for our experiments on this broad, standard concept drift testbench. This attests to the complexity that simple design decisions may produce in conjunction, leading to significant effects on algorithm performance---even in such simple artifacts as decision trees. Table \ref{table104} shows us the combined winning effect of these unspecified features for VFDT in the case of streams with concept drift.

\begin{table}[!ht]
	
	\caption{\label{table104}All three unspecified strategies put together}
	\scriptsize
	\centering
	\makebox[0.6\textwidth]{
		
		\begin{tabular}{m{12cm}?m{1.4cm}?m{1.4cm}}
			\toprule
			\multicolumn{1}{c?}{\textbf{Streams}} &  \multicolumn{1}{m{1.4cm}}{\textbf{VFDT}} &
			\textbf{VFDT: unspecified strategies}\\
			\midrule
			RecurrentConceptDriftStream -x 200000 -y 200000 -z 100 -s (AgrawalGenerator -f 2 -i 2) -d (AgrawalGenerator -f 3 -i 3) & 0.20846 & \textbf{0.1977} \\
RecurrentConceptDriftStream -x 200000 -y 200000 -z 100 -s (RandomTreeGenerator -r 1 -i 1) -d (RandomTreeGenerator -r 2 -i 2) & 0.22404 & \textbf{0.22123} \\
RecurrentConceptDriftStream -x 200000 -y 200000 -z 100 -s (SEAGenerator -f 2 -i 2) -d (SEAGenerator -f 3 -i 3) & 0.15251 & \textbf{0.15237} \\
RecurrentConceptDriftStream -x 200000 -y 200000 -z 100 -s (STAGGERGenerator -i 2 -f 2) -d (STAGGERGenerator -i 3 -f 3) & 0.1882 & \textbf{0.00699} \\
HyperplaneGenerator -k 10 -t 0.0001 -i 2 & 0.11566 & \textbf{0.11464} \\
HyperplaneGenerator -k 10 -t 0.001 -i 2 & 0.16785 & \textbf{0.16436} \\
HyperplaneGenerator -k 10 -t 0.01 -i 2 & 0.1713 & \textbf{0.16945} \\
HyperplaneGenerator -k 5 -t 0.0001 -i 2 & 0.1074 & \textbf{0.10723} \\
HyperplaneGenerator -k 5 -t 0.001 -i 2 & 0.16309 & \textbf{0.15871} \\
HyperplaneGenerator -k 5 -t 0.01 -i 2 & 0.17108 & \textbf{0.16479} \\
LEDGeneratorDrift -d 1 -i 2 & 0.26093 & \textbf{0.26087} \\
LEDGeneratorDrift -d 3 -i 2 & 0.26093 & \textbf{0.26087} \\
LEDGeneratorDrift -d 5 -i 2 & 0.26093 & \textbf{0.26087} \\
LEDGeneratorDrift -d 7 -i 2 & 0.26093 & \textbf{0.26087} \\
RandomRBFGeneratorDrift -s 0.0001 -k 10 -i 2 -r 2 & 0.11462 & \textbf{0.11331} \\
RandomRBFGeneratorDrift -s 0.0001 -k 50 -i 2 -r 2 & 0.2858 & \textbf{0.28356} \\
RandomRBFGeneratorDrift -s 0.001 -k 10 -i 2 -r 2 & 0.13821 & \textbf{0.13803} \\
RandomRBFGeneratorDrift -s 0.001 -k 50 -i 2 -r 2 & 0.40874 & \textbf{0.40593} \\
WaveformGeneratorDrift -d 1 -i 2 -n & 0.16284 & \textbf{0.16271} \\
WaveformGeneratorDrift -d 3 -i 2 -n & 0.16284 & \textbf{0.16271} \\
WaveformGeneratorDrift -d 5 -i 2 -n & 0.16284 & \textbf{0.16271} \\
WaveformGeneratorDrift -d 7 -i 2 -n & 0.16284 & \textbf{0.16271} \\
AbruptDriftGenerator -c  -o 1.0 -z 2 -n 2 -v 2 -r 2 -b 200000 -d Recurrent & 0.35403 & \textbf{0.00263} \\
AbruptDriftGenerator -c  -o 1.0 -z 3 -n 2 -v 2 -r 2 -b 200000 -d Recurrent & 0.37862 & \textbf{0.00328} \\
AbruptDriftGenerator -c  -o 1.0 -z 3 -n 3 -v 2 -r 2 -b 200000 -d Recurrent & 0.3504 & \textbf{0.01349} \\
AbruptDriftGenerator -c  -o 1.0 -z 3 -n 3 -v 3 -r 2 -b 200000 -d Recurrent & 0.36505 & \textbf{0.05837} \\
AbruptDriftGenerator -c  -o 1.0 -z 3 -n 3 -v 4 -r 2 -b 200000 -d Recurrent & 0.39687 & \textbf{0.15038} \\
AbruptDriftGenerator -c  -o 1.0 -z 3 -n 3 -v 5 -r 2 -b 200000 -d Recurrent & 0.39622 & \textbf{0.23056} \\
AbruptDriftGenerator -c  -o 1.0 -z 4 -n 2 -v 2 -r 2 -b 200000 -d Recurrent & 0.33416 & \textbf{0.0096} \\
AbruptDriftGenerator -c  -o 1.0 -z 4 -n 4 -v 4 -r 2 -b 200000 -d Recurrent & 0.40671 & \textbf{0.34814} \\
AbruptDriftGenerator -c  -o 1.0 -z 5 -n 2 -v 2 -r 2 -b 200000 -d Recurrent & 0.3309 & \textbf{0.00948} \\
AbruptDriftGenerator -c  -o 1.0 -z 5 -n 5 -v 5 -r 2 -b 200000 -d Recurrent & 0.46461 & \textbf{0.4605} \\
\bottomrule &  &  \\
\begin{tabularx}{\linewidth}{Xr}
		A \textbf{bold} value indicates higher accuracy, and \textit{\textbf{bold italics}} indicate a tie.  & \textbf{Unique Wins}
		\end{tabularx} & \textbf{0} & \textbf{32} \\
\cmidrule[0.4pt](lr){2-3} &  &  \\
\begin{tabularx}{\linewidth}{Xr} The test is a one-tailed binomial test to determine the probability that the strategy in the  
             rightmost column would achieve so many wins if wins and losses were equiprobable. & \textbf{Test Statistics} \end{tabularx} & \textbf{p-value: is $<$ 0.00001} & \textbf{Confidence Interval:  0.91063 --- 1} \\

			\bottomrule
		\end{tabular}
	}
\end{table}

\subsection{The Node Evisceration Strategy}

Given the success of the unspecified inadvertent strategy of resplitting described in Section \ref{ht-resplitting}, we experimented with a deliberate policy of eviscerating nodes instead, that is, when a previously used attribute resurfaced as the best, we cleared both node statistics and the class distribution within the node instead of allowing redundant ``re-splitting'' of the node on a used attribute.  Table \ref{table105} shows us that this deliberate policy works out to be immensely successful and would be worth trying for users of online decision trees.

\label{ht-evisceration}
\begin{table}[!ht]
	
	\caption{\label{table105}Clearing node statistics and Class distributions}
	\scriptsize
	\centering
	\makebox[0.6\textwidth]{
		
		\begin{tabular}{m{12cm}?m{1.4cm}?m{1.4cm}}
			\toprule
			\multicolumn{1}{c?}{\textbf{Streams}} &  \multicolumn{1}{m{1.4cm}}{\textbf{VFDT}} &
			\textbf{VFDT with deliberate node clearing}\\
			\midrule
			RecurrentConceptDriftStream -x 200000 -y 200000 -z 100 -s (AgrawalGenerator -f 2 -i 2) -d (AgrawalGenerator -f 3 -i 3) & 0.20846 & \textbf{0.17856} \\
RecurrentConceptDriftStream -x 200000 -y 200000 -z 100 -s (RandomTreeGenerator -r 1 -i 1) -d (RandomTreeGenerator -r 2 -i 2) & 0.22404 & \textbf{0.22004} \\
RecurrentConceptDriftStream -x 200000 -y 200000 -z 100 -s (SEAGenerator -f 2 -i 2) -d (SEAGenerator -f 3 -i 3) & 0.15251 & \textbf{0.14994} \\
RecurrentConceptDriftStream -x 200000 -y 200000 -z 100 -s (STAGGERGenerator -i 2 -f 2) -d (STAGGERGenerator -i 3 -f 3) & 0.1882 & \textbf{0.00435} \\
HyperplaneGenerator -k 10 -t 0.0001 -i 2 & 0.11566 & \textbf{0.11429} \\
HyperplaneGenerator -k 10 -t 0.001 -i 2 & 0.16785 & \textbf{0.16007} \\
HyperplaneGenerator -k 10 -t 0.01 -i 2 & 0.1713 & \textbf{0.1639} \\
HyperplaneGenerator -k 5 -t 0.0001 -i 2 & 0.1074 & \textbf{0.10719} \\
HyperplaneGenerator -k 5 -t 0.001 -i 2 & 0.16309 & \textbf{0.15347} \\
HyperplaneGenerator -k 5 -t 0.01 -i 2 & 0.17108 & \textbf{0.15901} \\
LEDGeneratorDrift -d 1 -i 2 & \textit{\textbf{0.26093}} & \textit{\textbf{0.26093}} \\
LEDGeneratorDrift -d 3 -i 2 & \textit{\textbf{0.26093}} & \textit{\textbf{0.26093}} \\
LEDGeneratorDrift -d 5 -i 2 & \textit{\textbf{0.26093}} & \textit{\textbf{0.26093}} \\
LEDGeneratorDrift -d 7 -i 2 & \textit{\textbf{0.26093}} & \textit{\textbf{0.26093}} \\
RandomRBFGeneratorDrift -s 0.0001 -k 10 -i 2 -r 2 & 0.11462 & \textbf{0.11451} \\
RandomRBFGeneratorDrift -s 0.0001 -k 50 -i 2 -r 2 & 0.2858 & \textbf{0.28429} \\
RandomRBFGeneratorDrift -s 0.001 -k 10 -i 2 -r 2 & \textbf{0.13821} & 0.13853 \\
RandomRBFGeneratorDrift -s 0.001 -k 50 -i 2 -r 2 & 0.40874 & \textbf{0.40525} \\
WaveformGeneratorDrift -d 1 -i 2 -n & \textit{\textbf{0.16284}} & \textit{\textbf{0.16284}} \\
WaveformGeneratorDrift -d 3 -i 2 -n & \textit{\textbf{0.16284}} & \textit{\textbf{0.16284}} \\
WaveformGeneratorDrift -d 5 -i 2 -n & \textit{\textbf{0.16284}} & \textit{\textbf{0.16284}} \\
WaveformGeneratorDrift -d 7 -i 2 -n & \textit{\textbf{0.16284}} & \textit{\textbf{0.16284}} \\
AbruptDriftGenerator -c  -o 1.0 -z 2 -n 2 -v 2 -r 2 -b 200000 -d Recurrent & 0.35403 & \textbf{0.00233} \\
AbruptDriftGenerator -c  -o 1.0 -z 3 -n 2 -v 2 -r 2 -b 200000 -d Recurrent & 0.37862 & \textbf{0.00344} \\
AbruptDriftGenerator -c  -o 1.0 -z 3 -n 3 -v 2 -r 2 -b 200000 -d Recurrent & 0.3504 & \textbf{0.01317} \\
AbruptDriftGenerator -c  -o 1.0 -z 3 -n 3 -v 3 -r 2 -b 200000 -d Recurrent & 0.36505 & \textbf{0.05836} \\
AbruptDriftGenerator -c  -o 1.0 -z 3 -n 3 -v 4 -r 2 -b 200000 -d Recurrent & 0.39687 & \textbf{0.15157} \\
AbruptDriftGenerator -c  -o 1.0 -z 3 -n 3 -v 5 -r 2 -b 200000 -d Recurrent & 0.39622 & \textbf{0.23064} \\
AbruptDriftGenerator -c  -o 1.0 -z 4 -n 2 -v 2 -r 2 -b 200000 -d Recurrent & 0.33416 & \textbf{0.00933} \\
AbruptDriftGenerator -c  -o 1.0 -z 4 -n 4 -v 4 -r 2 -b 200000 -d Recurrent & 0.40671 & \textbf{0.34795} \\
AbruptDriftGenerator -c  -o 1.0 -z 5 -n 2 -v 2 -r 2 -b 200000 -d Recurrent & 0.3309 & \textbf{0.00972} \\
AbruptDriftGenerator -c  -o 1.0 -z 5 -n 5 -v 5 -r 2 -b 200000 -d Recurrent & 0.46461 & \textbf{0.46356} \\
\bottomrule &  &  \\
\begin{tabularx}{\linewidth}{Xr}
		A \textbf{bold} value indicates higher accuracy, and \textit{\textbf{bold italics}} indicate a tie.  & \textbf{Unique Wins}
		\end{tabularx} & \textbf{1} & \textbf{23} \\
\cmidrule[0.4pt](lr){2-3} &  &  \\
\begin{tabularx}{\linewidth}{Xr} The test is a one-tailed binomial test to determine the probability that the strategy in the  
             column is likely to have an equal number of ``wins'' as the strategy being compared against on the left. & \textbf{Test Statistics} \end{tabularx} & \textbf{p-value: is $<$ 0.00001} & \textbf{Confidence Interval:  0.81711 --- 1} \\

			\bottomrule
		\end{tabular}
	}
\end{table}

\section{Unspecified and Emergent Behaviors in the MOA implementation of Hoeffding Adaptive Tree}
\label{hat}

We also investigate the success of Hoeffding Adaptive Tree (HAT)---the state-of-the-art adaptive tree learner---for learning under concept drift. We show that in addition to its main proposed strategy of building alternate subtrees upon drift detection and replacing mainline subtrees with alternates with lower error, HAT's excellent performance is influenced, to various degrees, by factors such as: 

\renewcommand{\labelitemi}{\scriptsize$\bullet$}
\begin{itemize}
	\item Alternate voting, explained in Sections \ref{hat-alt-vote}, \ref{hat-alt-of-alt}, and \ref{hat-alt-of-alt-no-single-leaves}
	\item Partial weighting of examples at leaves, explained in Section \ref{hat-partial-weighting}
	\item Effects from the VFDT base---explained in Sections \ref{hat-resplitting}, \ref{hat-infogainavg}, \ref{hat-ht-effects}
\end{itemize}

We elaborate on these mechanisms, perform comparisons with and without unspecified strategies enabled, and show the effects each strategy has on prequential error. We also describe unspecified features that do not result in significant effects.

\subsection{Resplitting on nominal attributes}
\label{hat-resplitting}

This unspecified feature of MOA-VFDT is responsible for a significant performance gain for HAT. As explained in Section \ref{ht-infogain}, nominal attributes that have been used once to split are ``reused", in that a single child is created if a previously used attribute leads to greater Information Gain than any other attribute under drifting streams. This turns out to be a particularly effective performance booster, as it forces the creation of a fresh leaf with empty node statistics, implying that every iteration of self-splitting leads to telescoping amnesia in the node statistics. We need the node statistics to compute Information Gain; when these are set to zero, we are effectively working with a blank slate, having deleted the historical node statistics that would otherwise retain information from older concepts. In addition, further splits will also have a predictor---a class distribution---derived solely from the freshly cleared node statistics of the parent that represent the post-drift concept. This enables learning a newer concept rapidly and is useful particularly in concept drift settings, more so for HAT than for VFDT, as is clear from our results. Table \ref{table6}.

\begin{table}[!ht]
	
	\caption{\label{table6}Performance of the resplitting on nominal attributes strategy
	} 
	\scriptsize
	\centering
	\makebox[0.6\textwidth]{
		
		\begin{tabular}{m{12cm}?m{1.4cm}?m{1.4cm}}
			\toprule
			\multicolumn{1}{c?}{\textbf{Streams}} &  \multicolumn{1}{m{1.4cm}}{\textbf{HAT that does not ``resplit'' on nominal attributes}} &
			\textbf{HAT that ``resplits'' on nominal attributes}\\
			\midrule
			RecurrentConceptDriftStream -x 200000 -y 200000 -z 100 -s (AgrawalGenerator -f 2 -i 2) -d (AgrawalGenerator -f 3 -i 3) & 0.13645 & \textbf{0.12454} \\
RecurrentConceptDriftStream -x 200000 -y 200000 -z 100 -s (RandomTreeGenerator -r 1 -i 1) -d (RandomTreeGenerator -r 2 -i 2) & 0.10104 & \textbf{0.09866} \\
RecurrentConceptDriftStream -x 200000 -y 200000 -z 100 -s (SEAGenerator -f 2 -i 2) -d (SEAGenerator -f 3 -i 3) & 0.12579 & \textbf{0.11207} \\
RecurrentConceptDriftStream -x 200000 -y 200000 -z 100 -s (STAGGERGenerator -i 2 -f 2) -d (STAGGERGenerator -i 3 -f 3) & 0.00215 & \textbf{0.00213} \\
HyperplaneGenerator -k 10 -t 0.0001 -i 2 & \textbf{0.10801} & 0.10875 \\
HyperplaneGenerator -k 10 -t 0.001 -i 2 & 0.11934 & \textbf{0.11823} \\
HyperplaneGenerator -k 10 -t 0.01 -i 2 & 0.1198 & \textbf{0.11948} \\
HyperplaneGenerator -k 5 -t 0.0001 -i 2 & \textbf{0.10588} & 0.10693 \\
HyperplaneGenerator -k 5 -t 0.001 -i 2 & 0.11254 & \textbf{0.11008} \\
HyperplaneGenerator -k 5 -t 0.01 -i 2 & 0.10767 & \textbf{0.1049} \\
LEDGeneratorDrift -d 1 -i 2 & \textit{\textbf{0.26137}} & \textit{\textbf{0.26137}} \\
LEDGeneratorDrift -d 3 -i 2 & \textit{\textbf{0.26137}} & \textit{\textbf{0.26137}} \\
LEDGeneratorDrift -d 5 -i 2 & \textit{\textbf{0.26137}} & \textit{\textbf{0.26137}} \\
LEDGeneratorDrift -d 7 -i 2 & \textit{\textbf{0.26137}} & \textit{\textbf{0.26137}} \\
RandomRBFGeneratorDrift -s 0.0001 -k 10 -i 2 -r 2 & 0.13508 & \textbf{0.1197} \\
RandomRBFGeneratorDrift -s 0.0001 -k 50 -i 2 -r 2 & 0.19549 & \textbf{0.19199} \\
RandomRBFGeneratorDrift -s 0.001 -k 10 -i 2 -r 2 & 0.16636 & \textbf{0.15267} \\
RandomRBFGeneratorDrift -s 0.001 -k 50 -i 2 -r 2 & 0.33926 & \textbf{0.33877} \\
WaveformGeneratorDrift -d 1 -i 2 -n & \textbf{0.16279} & 0.16337 \\
WaveformGeneratorDrift -d 3 -i 2 -n & \textbf{0.16279} & 0.16337 \\
WaveformGeneratorDrift -d 5 -i 2 -n & \textbf{0.16279} & 0.16337 \\
WaveformGeneratorDrift -d 7 -i 2 -n & \textbf{0.16279} & 0.16337 \\
AbruptDriftGenerator -c  -o 1.0 -z 2 -n 2 -v 2 -r 2 -b 200000 -d Recurrent & 0.00185 & \textbf{0.00085} \\
AbruptDriftGenerator -c  -o 1.0 -z 3 -n 2 -v 2 -r 2 -b 200000 -d Recurrent & 0.0016 & \textbf{0.00087} \\
AbruptDriftGenerator -c  -o 1.0 -z 3 -n 3 -v 2 -r 2 -b 200000 -d Recurrent & 0.00252 & \textbf{0.00169} \\
AbruptDriftGenerator -c  -o 1.0 -z 3 -n 3 -v 3 -r 2 -b 200000 -d Recurrent & 0.00425 & \textbf{0.00404} \\
AbruptDriftGenerator -c  -o 1.0 -z 3 -n 3 -v 4 -r 2 -b 200000 -d Recurrent & \textbf{0.01212} & 0.01228 \\
AbruptDriftGenerator -c  -o 1.0 -z 3 -n 3 -v 5 -r 2 -b 200000 -d Recurrent & 0.01687 & \textbf{0.01675} \\
AbruptDriftGenerator -c  -o 1.0 -z 4 -n 2 -v 2 -r 2 -b 200000 -d Recurrent & 0.00177 & \textbf{0.00095} \\
AbruptDriftGenerator -c  -o 1.0 -z 4 -n 4 -v 4 -r 2 -b 200000 -d Recurrent & 0.06662 & \textbf{0.06643} \\
AbruptDriftGenerator -c  -o 1.0 -z 5 -n 2 -v 2 -r 2 -b 200000 -d Recurrent & 0.00177 & \textbf{0.00113} \\
AbruptDriftGenerator -c  -o 1.0 -z 5 -n 5 -v 5 -r 2 -b 200000 -d Recurrent & \textit{\textbf{0.34892}} & \textit{\textbf{0.34892}} \\
\bottomrule &  &  \\
\begin{tabularx}{\linewidth}{Xr}
		A \textbf{bold} value indicates higher accuracy, and \textit{\textbf{bold italics}} indicate a tie.  & \textbf{Unique Wins}
		\end{tabularx} & \textbf{7} & \textbf{20} \\
\cmidrule[0.4pt](lr){2-3} &  &  \\
\begin{tabularx}{\linewidth}{Xr} The test is a one-tailed binomial test to determine the probability that the strategy in the  
             rightmost column would achieve so many wins if wins and losses were equiprobable. & \textbf{Test Statistics} \end{tabularx} & \textbf{p-value: 0.00958} & \textbf{Confidence Interval:  0.5677 --- 1} \\

			\bottomrule
		\end{tabular}
	}
\end{table}

\subsection{Alternate Voting}
\label{hat-alt-vote}

\begin{table}[!ht]
	
	\caption{\label{table1}Performance of the single alternate voting strategy: a single alternate may be grown and allowed to vote.\\
	While the base algorithm makes no provision for alternates voting, allowing them to do so provides the system with a form of lookahead wherein alternate subtrees under development---concepts being developed as potential future replacements---begin to take part in providing predictions instead of waiting until a replacement has taken place to start doing so.} 
	\scriptsize
	\centering
	\makebox[0.6\textwidth]{
		
		\begin{tabular}{m{12cm}?m{1.4cm}?m{1.4cm}}
			\toprule
			\multicolumn{1}{c?}{\textbf{Streams}} &  \multicolumn{1}{m{1.4cm}}{\textbf{HAT}} &
			\textbf{HAT with a single voting alternate}\\
			\midrule
			RecurrentConceptDriftStream -x 200000 -y 200000 -z 100 -s (AgrawalGenerator -f 2 -i 2) -d (AgrawalGenerator -f 3 -i 3) & 0.13645 & \textbf{0.13643} \\
RecurrentConceptDriftStream -x 200000 -y 200000 -z 100 -s (RandomTreeGenerator -r 1 -i 1) -d (RandomTreeGenerator -r 2 -i 2) & \textbf{0.10104} & 0.10615 \\
RecurrentConceptDriftStream -x 200000 -y 200000 -z 100 -s (SEAGenerator -f 2 -i 2) -d (SEAGenerator -f 3 -i 3) & 0.12579 & \textbf{0.12538} \\
RecurrentConceptDriftStream -x 200000 -y 200000 -z 100 -s (STAGGERGenerator -i 2 -f 2) -d (STAGGERGenerator -i 3 -f 3) & 0.00215 & \textbf{0.00136} \\
HyperplaneGenerator -k 10 -t 0.0001 -i 2 & 0.10801 & \textbf{0.10576} \\
HyperplaneGenerator -k 10 -t 0.001 -i 2 & 0.11934 & \textbf{0.11606} \\
HyperplaneGenerator -k 10 -t 0.01 -i 2 & 0.1198 & \textbf{0.11681} \\
HyperplaneGenerator -k 5 -t 0.0001 -i 2 & 0.10588 & \textbf{0.10412} \\
HyperplaneGenerator -k 5 -t 0.001 -i 2 & 0.11254 & \textbf{0.10947} \\
HyperplaneGenerator -k 5 -t 0.01 -i 2 & 0.10767 & \textbf{0.1047} \\
LEDGeneratorDrift -d 1 -i 2 & 0.26137 & \textbf{0.26123} \\
LEDGeneratorDrift -d 3 -i 2 & 0.26137 & \textbf{0.26123} \\
LEDGeneratorDrift -d 5 -i 2 & 0.26137 & \textbf{0.26123} \\
LEDGeneratorDrift -d 7 -i 2 & 0.26137 & \textbf{0.26123} \\
RandomRBFGeneratorDrift -s 0.0001 -k 10 -i 2 -r 2 & 0.13508 & \textbf{0.13109} \\
RandomRBFGeneratorDrift -s 0.0001 -k 50 -i 2 -r 2 & 0.19549 & \textbf{0.18094} \\
RandomRBFGeneratorDrift -s 0.001 -k 10 -i 2 -r 2 & 0.16636 & \textbf{0.16503} \\
RandomRBFGeneratorDrift -s 0.001 -k 50 -i 2 -r 2 & 0.33926 & \textbf{0.32535} \\
WaveformGeneratorDrift -d 1 -i 2 -n & 0.16279 & \textbf{0.16251} \\
WaveformGeneratorDrift -d 3 -i 2 -n & 0.16279 & \textbf{0.16251} \\
WaveformGeneratorDrift -d 5 -i 2 -n & 0.16279 & \textbf{0.16251} \\
WaveformGeneratorDrift -d 7 -i 2 -n & 0.16279 & \textbf{0.16251} \\
AbruptDriftGenerator -c  -o 1.0 -z 2 -n 2 -v 2 -r 2 -b 200000 -d Recurrent & 0.00185 & \textbf{7e-04} \\
AbruptDriftGenerator -c  -o 1.0 -z 3 -n 2 -v 2 -r 2 -b 200000 -d Recurrent & 0.0016 & \textbf{0.00056} \\
AbruptDriftGenerator -c  -o 1.0 -z 3 -n 3 -v 2 -r 2 -b 200000 -d Recurrent & 0.00252 & \textbf{0.00144} \\
AbruptDriftGenerator -c  -o 1.0 -z 3 -n 3 -v 3 -r 2 -b 200000 -d Recurrent & \textbf{0.00425} & 0.00475 \\
AbruptDriftGenerator -c  -o 1.0 -z 3 -n 3 -v 4 -r 2 -b 200000 -d Recurrent & \textbf{0.01212} & 0.01389 \\
AbruptDriftGenerator -c  -o 1.0 -z 3 -n 3 -v 5 -r 2 -b 200000 -d Recurrent & \textbf{0.01687} & 0.01978 \\
AbruptDriftGenerator -c  -o 1.0 -z 4 -n 2 -v 2 -r 2 -b 200000 -d Recurrent & 0.00177 & \textbf{0.00049} \\
AbruptDriftGenerator -c  -o 1.0 -z 4 -n 4 -v 4 -r 2 -b 200000 -d Recurrent & \textbf{0.06662} & 0.07347 \\
AbruptDriftGenerator -c  -o 1.0 -z 5 -n 2 -v 2 -r 2 -b 200000 -d Recurrent & 0.00177 & \textbf{0.00037} \\
AbruptDriftGenerator -c  -o 1.0 -z 5 -n 5 -v 5 -r 2 -b 200000 -d Recurrent & \textbf{0.34892} & 0.35605 \\
\bottomrule &  &  \\
\begin{tabularx}{\linewidth}{Xr}
		A \textbf{bold} value indicates higher accuracy, and \textit{\textbf{bold italics}} indicate a tie.  & \textbf{Unique Wins}
		\end{tabularx} & \textbf{6} & \textbf{26} \\
\cmidrule[0.4pt](lr){2-3} &  &  \\
\begin{tabularx}{\linewidth}{Xr} The test is a one-tailed binomial test to determine the probability that the strategy in the  
             rightmost column would achieve so many wins if wins and losses were equiprobable. & \textbf{Test Statistics} \end{tabularx} & \textbf{p-value: 0.00027} & \textbf{Confidence Interval:  0.66313 --- 1} \\

			\bottomrule
		\end{tabular}
	}
\end{table}

HAT makes use of change detectors at each node to determine whether a concept drift has occurred. Upon detecting a concept drift, an alternate subtree is grown. After a set period, the alternate subtree is made eligible to replace the mainline subtree if it has lower prequential error. Alternates are only specified to be potential replacements, not to vote. After all, it is possible that an alternate is grown as a result of a false drift detection due to noise.

However, should alternates be allowed to vote, they can represent a lookahead ability; while the model currently holds on to an older concept due to underlying statistical considerations of the HoeffdingTree base, exploratory submodels that catch on to fresh concepts may represent the best future state of the tree in concordance with ongoing concept drift. Giving these alternates a say in model predictions would be dependent on the scenario: it is a tradeoff between identifying subtrees that represent new concepts early and allowing in predictions from subtrees built as a result of noise.

Alternates voting was one of the unspecified features in the original implementation of HAT. We have distilled this feature in order to study it in isolation.

Table \ref{table1} compares a baseline, standardized HAT (without the HoeffdingTree unspecified effects described in Sections \ref{ht-resplitting}, \ref{ht-infogain}, and \ref{ht-nodetime} or the other unspecified HAT features described in this section), with a version of HAT where only a single alternate is grown and \textit{allowed to vote}. 

It is clear from Table \ref{table1} that allowing alternates to vote is beneficial for prequential accuracy on a standard testbench of drifting streams; we also notice a clear trend where increasing dimensionality of the data leads to loss of performance for the alternate voting strategy (as seen in the last ten rows).

The base strategy wins on streams with greater data dimensionality. The recurrent abrupt drift streams are notated with the number of attributes, followed by the number of values per attribute, and the number of classes; a 5x4x3 stream has 5 nominal attributes, 4 values per attribute, and 3 possible classes. The experimental results in Table \ref{table1} show us that the base HAT strategy wins on 5x5x5, 4x4x4, 3x3x5, 3x3x4, and 3x3x3 streams, and the Random Tree Generator stream; in this test setting, it loses on those streams with under 21 degrees of freedom. 

Given that the high-dimensional recurrent drift streams are noise free, we hypothesize that the effect due to alternate voting---an aggressive lookahead strategy---results in higher bias (implying a larger bias space or set of hypothesis functions to choose from). In increasingly higher dimensional scenarios, the larger bias---and the ability to traverse it thanks to model adaptation---would lead to overfitting recent examples and thus a drop in prequential accuracy performance.

Without the results on high dimensional streams, the alternate voting strategy never loses! Thus on a typical concept drift testbench found in the literature that does not include a high dimensional recurrent abrupt drift generator, the alternate voting strategy leads to almost universal outperformance over standard HAT. It might be interesting to explore the extent to which existing results on the standard testbench generalize to higher dimensionality.

Our conclusion with respect to unspecified alternate voting is that it has a profound, fundamental impact on the performance of HAT, in a positive sense on a standard concept drift testbench.

It should be noted that some of the results look identical; this is likely an issue with the lack of model diversity in those generating models in spite of parametrization. In particular, LED and Waveform drift streams show little variance in learner performance.

\subsection{Alternates of alternates}
\label{hat-alt-of-alt}

Table \ref{table2}: In addition to the losses on higher dimensional streams, enabling alternates to sprout their own alternates leads to losses on both the RBF parametrizations with fewer (10) centroids, irrespective of drift rate (0.001, 0.0001). We offer this observation without explanation; it merits further study. Overall, our conclusion based on this testbench is that this unspecified strategy does not provide improvement in prequential accuracy.

\begin{table}[!ht]
	
	\caption{\label{table2}Performance of the multiple alternates voting strategy: multiple alternates may be grown and allowed to vote} 
	\scriptsize
	\centering
	\makebox[0.6\textwidth]{
		
		\begin{tabular}{m{12cm}?m{1.4cm}?m{1.4cm}}
			\toprule
			\multicolumn{1}{c?}{\textbf{Streams}} &  \multicolumn{1}{m{1.4cm}}{\textbf{HAT with multiple voting alternates}} &
			\textbf{HAT with single voting alternate}\\
			\midrule
			RecurrentConceptDriftStream -x 200000 -y 200000 -z 100 -s (AgrawalGenerator -f 2 -i 2) -d (AgrawalGenerator -f 3 -i 3) & \textit{\textbf{0.13643}} & \textit{\textbf{0.13643}} \\
RecurrentConceptDriftStream -x 200000 -y 200000 -z 100 -s (RandomTreeGenerator -r 1 -i 1) -d (RandomTreeGenerator -r 2 -i 2) & 0.10663 & \textbf{0.10615} \\
RecurrentConceptDriftStream -x 200000 -y 200000 -z 100 -s (SEAGenerator -f 2 -i 2) -d (SEAGenerator -f 3 -i 3) & \textit{\textbf{0.12538}} & \textit{\textbf{0.12538}} \\
RecurrentConceptDriftStream -x 200000 -y 200000 -z 100 -s (STAGGERGenerator -i 2 -f 2) -d (STAGGERGenerator -i 3 -f 3) & \textit{\textbf{0.00136}} & \textit{\textbf{0.00136}} \\
HyperplaneGenerator -k 10 -t 0.0001 -i 2 & \textbf{0.10562} & 0.10576 \\
HyperplaneGenerator -k 10 -t 0.001 -i 2 & \textbf{0.11571} & 0.11606 \\
HyperplaneGenerator -k 10 -t 0.01 -i 2 & 0.11705 & \textbf{0.11681} \\
HyperplaneGenerator -k 5 -t 0.0001 -i 2 & 0.1044 & \textbf{0.10412} \\
HyperplaneGenerator -k 5 -t 0.001 -i 2 & 0.10963 & \textbf{0.10947} \\
HyperplaneGenerator -k 5 -t 0.01 -i 2 & \textbf{0.10469} & 0.1047 \\
LEDGeneratorDrift -d 1 -i 2 & 0.26127 & \textbf{0.26123} \\
LEDGeneratorDrift -d 3 -i 2 & 0.26127 & \textbf{0.26123} \\
LEDGeneratorDrift -d 5 -i 2 & 0.26127 & \textbf{0.26123} \\
LEDGeneratorDrift -d 7 -i 2 & 0.26127 & \textbf{0.26123} \\
RandomRBFGeneratorDrift -s 0.0001 -k 10 -i 2 -r 2 & 0.13617 & \textbf{0.13109} \\
RandomRBFGeneratorDrift -s 0.0001 -k 50 -i 2 -r 2 & 0.18238 & \textbf{0.18094} \\
RandomRBFGeneratorDrift -s 0.001 -k 10 -i 2 -r 2 & 0.17105 & \textbf{0.16503} \\
RandomRBFGeneratorDrift -s 0.001 -k 50 -i 2 -r 2 & \textbf{0.32534} & 0.32535 \\
WaveformGeneratorDrift -d 1 -i 2 -n & \textbf{0.16234} & 0.16251 \\
WaveformGeneratorDrift -d 3 -i 2 -n & \textbf{0.16234} & 0.16251 \\
WaveformGeneratorDrift -d 5 -i 2 -n & \textbf{0.16234} & 0.16251 \\
WaveformGeneratorDrift -d 7 -i 2 -n & \textbf{0.16234} & 0.16251 \\
AbruptDriftGenerator -c  -o 1.0 -z 2 -n 2 -v 2 -r 2 -b 200000 -d Recurrent & \textit{\textbf{7e-04}} & \textit{\textbf{7e-04}} \\
AbruptDriftGenerator -c  -o 1.0 -z 3 -n 2 -v 2 -r 2 -b 200000 -d Recurrent & \textit{\textbf{0.00056}} & \textit{\textbf{0.00056}} \\
AbruptDriftGenerator -c  -o 1.0 -z 3 -n 3 -v 2 -r 2 -b 200000 -d Recurrent & 0.00157 & \textbf{0.00144} \\
AbruptDriftGenerator -c  -o 1.0 -z 3 -n 3 -v 3 -r 2 -b 200000 -d Recurrent & 0.00793 & \textbf{0.00475} \\
AbruptDriftGenerator -c  -o 1.0 -z 3 -n 3 -v 4 -r 2 -b 200000 -d Recurrent & 0.01693 & \textbf{0.01389} \\
AbruptDriftGenerator -c  -o 1.0 -z 3 -n 3 -v 5 -r 2 -b 200000 -d Recurrent & 0.0212 & \textbf{0.01978} \\
AbruptDriftGenerator -c  -o 1.0 -z 4 -n 2 -v 2 -r 2 -b 200000 -d Recurrent & \textit{\textbf{0.00049}} & \textit{\textbf{0.00049}} \\
AbruptDriftGenerator -c  -o 1.0 -z 4 -n 4 -v 4 -r 2 -b 200000 -d Recurrent & 0.07469 & \textbf{0.07347} \\
AbruptDriftGenerator -c  -o 1.0 -z 5 -n 2 -v 2 -r 2 -b 200000 -d Recurrent & \textit{\textbf{0.00037}} & \textit{\textbf{0.00037}} \\
AbruptDriftGenerator -c  -o 1.0 -z 5 -n 5 -v 5 -r 2 -b 200000 -d Recurrent & 0.35638 & \textbf{0.35605} \\
\bottomrule &  &  \\
\begin{tabularx}{\linewidth}{Xr}
		A \textbf{bold} value indicates higher accuracy, and \textit{\textbf{bold italics}} indicate a tie.  & \textbf{Unique Wins}
		\end{tabularx} & \textbf{8} & \textbf{17} \\
\cmidrule[0.4pt](lr){2-3} &  &  \\
\begin{tabularx}{\linewidth}{Xr} The test is a one-tailed binomial test to determine the probability that the strategy in the  
             rightmost column would achieve so many wins if wins and losses were equiprobable. & \textbf{Test Statistics} \end{tabularx} & \textbf{p-value: 0.05388} & \textbf{Confidence Interval:  0.49636 --- 1} \\

			\bottomrule
		\end{tabular}
	}
\end{table}

\subsection{Alternates of alternates allowed, but only single leaf alternates do not vote}
\label{hat-alt-of-alt-no-single-leaves}

This is the unspecified voting strategy followed in the implementation used to report results in the original paper; alternates are allowed to vote, and alternates may grow alternates, which are also allowed to vote---but only if the alternates are not simple leaves but have more structure. Table \ref{table3} demonstrates that, interestingly, this strategy registers fewer wins than when alternates are allowed to vote regardless of how simple their structure (which we have examined in Table \ref{table2}); relative performance on tree-based streams does not change, but the strategy underperforms across the board---on gradual drift from the Hyperplane generator, on LED, RBF and Waveform streams. The implication is that the lookahead provided by single-leaf alternates is valuable in drifting environments. Note again repetitiveness on the Waveform and LED streams.

\begin{table}[!ht]
	
	\caption{\label{table3}Performance of the mutiple alternates voting strategies: when single leaves are not allowed to vote, and when they are} 
	\scriptsize
	\centering
	\makebox[0.6\textwidth]{
		
		\begin{tabular}{m{12cm}?m{1.4cm}?m{1.4cm}}
			\toprule
			\multicolumn{1}{c?}{\textbf{Streams}} &  \multicolumn{1}{m{1.4cm}}{\textbf{HAT with multiple voting alternates, excepting single leaves}} &
			\textbf{HAT with multiple voting alternates, including single leaves}\\
			\midrule
			RecurrentConceptDriftStream -x 200000 -y 200000 -z 100 -s (AgrawalGenerator -f 2 -i 2) -d (AgrawalGenerator -f 3 -i 3) & \textbf{0.1364} & 0.13643 \\
RecurrentConceptDriftStream -x 200000 -y 200000 -z 100 -s (RandomTreeGenerator -r 1 -i 1) -d (RandomTreeGenerator -r 2 -i 2) & \textbf{0.10284} & 0.10663 \\
RecurrentConceptDriftStream -x 200000 -y 200000 -z 100 -s (SEAGenerator -f 2 -i 2) -d (SEAGenerator -f 3 -i 3) & 0.1256 & \textbf{0.12538} \\
RecurrentConceptDriftStream -x 200000 -y 200000 -z 100 -s (STAGGERGenerator -i 2 -f 2) -d (STAGGERGenerator -i 3 -f 3) & 0.002 & \textbf{0.00136} \\
HyperplaneGenerator -k 10 -t 0.0001 -i 2 & 0.13141 & \textbf{0.10562} \\
HyperplaneGenerator -k 10 -t 0.001 -i 2 & 0.12012 & \textbf{0.11571} \\
HyperplaneGenerator -k 10 -t 0.01 -i 2 & 0.12594 & \textbf{0.11705} \\
HyperplaneGenerator -k 5 -t 0.0001 -i 2 & 0.13913 & \textbf{0.1044} \\
HyperplaneGenerator -k 5 -t 0.001 -i 2 & 0.11189 & \textbf{0.10963} \\
HyperplaneGenerator -k 5 -t 0.01 -i 2 & 0.10566 & \textbf{0.10469} \\
LEDGeneratorDrift -d 1 -i 2 & 0.26138 & \textbf{0.26127} \\
LEDGeneratorDrift -d 3 -i 2 & 0.26138 & \textbf{0.26127} \\
LEDGeneratorDrift -d 5 -i 2 & 0.26138 & \textbf{0.26127} \\
LEDGeneratorDrift -d 7 -i 2 & 0.26138 & \textbf{0.26127} \\
RandomRBFGeneratorDrift -s 0.0001 -k 10 -i 2 -r 2 & 0.14943 & \textbf{0.13617} \\
RandomRBFGeneratorDrift -s 0.0001 -k 50 -i 2 -r 2 & 0.19643 & \textbf{0.18238} \\
RandomRBFGeneratorDrift -s 0.001 -k 10 -i 2 -r 2 & 0.20134 & \textbf{0.17105} \\
RandomRBFGeneratorDrift -s 0.001 -k 50 -i 2 -r 2 & 0.33891 & \textbf{0.32534} \\
WaveformGeneratorDrift -d 1 -i 2 -n & 0.16866 & \textbf{0.16234} \\
WaveformGeneratorDrift -d 3 -i 2 -n & 0.16866 & \textbf{0.16234} \\
WaveformGeneratorDrift -d 5 -i 2 -n & 0.16866 & \textbf{0.16234} \\
WaveformGeneratorDrift -d 7 -i 2 -n & 0.16866 & \textbf{0.16234} \\
AbruptDriftGenerator -c  -o 1.0 -z 2 -n 2 -v 2 -r 2 -b 200000 -d Recurrent & 0.00159 & \textbf{7e-04} \\
AbruptDriftGenerator -c  -o 1.0 -z 3 -n 2 -v 2 -r 2 -b 200000 -d Recurrent & 0.00147 & \textbf{0.00056} \\
AbruptDriftGenerator -c  -o 1.0 -z 3 -n 3 -v 2 -r 2 -b 200000 -d Recurrent & 0.00246 & \textbf{0.00157} \\
AbruptDriftGenerator -c  -o 1.0 -z 3 -n 3 -v 3 -r 2 -b 200000 -d Recurrent & \textbf{0.00699} & 0.00793 \\
AbruptDriftGenerator -c  -o 1.0 -z 3 -n 3 -v 4 -r 2 -b 200000 -d Recurrent & \textbf{0.01497} & 0.01693 \\
AbruptDriftGenerator -c  -o 1.0 -z 3 -n 3 -v 5 -r 2 -b 200000 -d Recurrent & \textbf{0.01796} & 0.0212 \\
AbruptDriftGenerator -c  -o 1.0 -z 4 -n 2 -v 2 -r 2 -b 200000 -d Recurrent & 0.00149 & \textbf{0.00049} \\
AbruptDriftGenerator -c  -o 1.0 -z 4 -n 4 -v 4 -r 2 -b 200000 -d Recurrent & \textbf{0.06767} & 0.07469 \\
AbruptDriftGenerator -c  -o 1.0 -z 5 -n 2 -v 2 -r 2 -b 200000 -d Recurrent & 0.00148 & \textbf{0.00037} \\
AbruptDriftGenerator -c  -o 1.0 -z 5 -n 5 -v 5 -r 2 -b 200000 -d Recurrent & \textbf{0.34955} & 0.35638 \\
\bottomrule &  &  \\
\begin{tabularx}{\linewidth}{Xr}
		A \textbf{bold} value indicates higher accuracy, and \textit{\textbf{bold italics}} indicate a tie.  & \textbf{Unique Wins}
		\end{tabularx} & \textbf{7} & \textbf{25} \\
\cmidrule[0.4pt](lr){2-3} &  &  \\
\begin{tabularx}{\linewidth}{Xr} The test is a one-tailed binomial test to determine the probability that the strategy in the  
             rightmost column would achieve so many wins if wins and losses were equiprobable. & \textbf{Test Statistics} \end{tabularx} & \textbf{p-value: 0.00105} & \textbf{Confidence Interval:  0.6281 --- 1} \\

			\bottomrule
		\end{tabular}
	}
\end{table}

\subsection{Partial weighting}
\label{hat-partial-weighting}

Weighting examples is a common strategy to force a learner to focus on some examples more so than others. For instance, online boosting \parencite{oza05} weights misclassified examples higher than correctly classified ones, creating a virtual ``heavier" history for examples that are being misclassified to aid learners in classifying them correctly if the reason for the misclassification happens to be sparsity of examples complicated by noisy observations.

Table \ref{table4} demonstrates that, with HAT, using a Poisson(1) weighting for examples at leaves reverses the losing trend due to only letting single leaf alternates vote (25 wins, up from 17 --- see Table \ref{table3} in Section \ref{hat-alt-of-alt-no-single-leaves}); some instances are weighted more than others at leaves, prioritising them. A Poisson(1) weighting assigns weights to examples stochastically---the weight is randomly drawn from the Poisson(1) distribution, with a mean value of 1, and otherwise taking nonnegative integer values. This strategy is followed in the original HAT implementation with the likely intention of creating a semblance of ensemble like diversity within subtrees. Given the seeming success of this unspecified strategy, it might be worthwhile exploring other Poisson values---Poisson(1) evaluates to 0 about a third of the time, so a third of all instances are not counted at leaves.

\begin{table}[!ht]
	
	\caption{\label{table4}Performance of the partial weighting strategy: leaves are weighted Poisson(1)
	} 
	\scriptsize
	\centering
	\makebox[0.6\textwidth]{
		
		\begin{tabular}{m{12cm}?m{1.4cm}?m{1.4cm}}
			\toprule
			\multicolumn{1}{c?}{\textbf{Streams}} &  \multicolumn{1}{m{1.4cm}}{\textbf{HAT with multiple voting alternates, excepting single leaves}} &
			\textbf{...With partial weighting enabled}\\
			\midrule
			RecurrentConceptDriftStream -x 200000 -y 200000 -z 100 -s (AgrawalGenerator -f 2 -i 2) -d (AgrawalGenerator -f 3 -i 3) & 0.1364 & \textbf{0.13354} \\
RecurrentConceptDriftStream -x 200000 -y 200000 -z 100 -s (RandomTreeGenerator -r 1 -i 1) -d (RandomTreeGenerator -r 2 -i 2) & 0.10284 & \textbf{0.10111} \\
RecurrentConceptDriftStream -x 200000 -y 200000 -z 100 -s (SEAGenerator -f 2 -i 2) -d (SEAGenerator -f 3 -i 3) & 0.1256 & \textbf{0.12522} \\
RecurrentConceptDriftStream -x 200000 -y 200000 -z 100 -s (STAGGERGenerator -i 2 -f 2) -d (STAGGERGenerator -i 3 -f 3) & \textbf{0.002} & 0.00205 \\
HyperplaneGenerator -k 10 -t 0.0001 -i 2 & \textbf{0.13141} & 0.15097 \\
HyperplaneGenerator -k 10 -t 0.001 -i 2 & \textbf{0.12012} & 0.14423 \\
HyperplaneGenerator -k 10 -t 0.01 -i 2 & 0.12594 & \textbf{0.12496} \\
HyperplaneGenerator -k 5 -t 0.0001 -i 2 & 0.13913 & \textbf{0.13817} \\
HyperplaneGenerator -k 5 -t 0.001 -i 2 & \textbf{0.11189} & 0.11276 \\
HyperplaneGenerator -k 5 -t 0.01 -i 2 & \textbf{0.10566} & 0.10773 \\
LEDGeneratorDrift -d 1 -i 2 & 0.26138 & \textbf{0.26114} \\
LEDGeneratorDrift -d 3 -i 2 & 0.26138 & \textbf{0.26114} \\
LEDGeneratorDrift -d 5 -i 2 & 0.26138 & \textbf{0.26114} \\
LEDGeneratorDrift -d 7 -i 2 & 0.26138 & \textbf{0.26114} \\
RandomRBFGeneratorDrift -s 0.0001 -k 10 -i 2 -r 2 & 0.14943 & \textbf{0.1475} \\
RandomRBFGeneratorDrift -s 0.0001 -k 50 -i 2 -r 2 & 0.19643 & \textbf{0.18406} \\
RandomRBFGeneratorDrift -s 0.001 -k 10 -i 2 -r 2 & 0.20134 & \textbf{0.19871} \\
RandomRBFGeneratorDrift -s 0.001 -k 50 -i 2 -r 2 & 0.33891 & \textbf{0.32724} \\
WaveformGeneratorDrift -d 1 -i 2 -n & \textbf{0.16866} & 0.1749 \\
WaveformGeneratorDrift -d 3 -i 2 -n & \textbf{0.16866} & 0.1749 \\
WaveformGeneratorDrift -d 5 -i 2 -n & \textbf{0.16866} & 0.1749 \\
WaveformGeneratorDrift -d 7 -i 2 -n & \textbf{0.16866} & 0.1749 \\
AbruptDriftGenerator -c  -o 1.0 -z 2 -n 2 -v 2 -r 2 -b 200000 -d Recurrent & 0.00159 & \textbf{0.00154} \\
AbruptDriftGenerator -c  -o 1.0 -z 3 -n 2 -v 2 -r 2 -b 200000 -d Recurrent & 0.00147 & \textbf{0.00135} \\
AbruptDriftGenerator -c  -o 1.0 -z 3 -n 3 -v 2 -r 2 -b 200000 -d Recurrent & 0.00246 & \textbf{0.00185} \\
AbruptDriftGenerator -c  -o 1.0 -z 3 -n 3 -v 3 -r 2 -b 200000 -d Recurrent & 0.00699 & \textbf{0.00698} \\
AbruptDriftGenerator -c  -o 1.0 -z 3 -n 3 -v 4 -r 2 -b 200000 -d Recurrent & 0.01497 & \textbf{0.01312} \\
AbruptDriftGenerator -c  -o 1.0 -z 3 -n 3 -v 5 -r 2 -b 200000 -d Recurrent & 0.01796 & \textbf{0.01528} \\
AbruptDriftGenerator -c  -o 1.0 -z 4 -n 2 -v 2 -r 2 -b 200000 -d Recurrent & 0.00149 & \textbf{0.00146} \\
AbruptDriftGenerator -c  -o 1.0 -z 4 -n 4 -v 4 -r 2 -b 200000 -d Recurrent & 0.06767 & \textbf{0.05249} \\
AbruptDriftGenerator -c  -o 1.0 -z 5 -n 2 -v 2 -r 2 -b 200000 -d Recurrent & 0.00148 & \textbf{0.00143} \\
AbruptDriftGenerator -c  -o 1.0 -z 5 -n 5 -v 5 -r 2 -b 200000 -d Recurrent & 0.34955 & \textbf{0.28932} \\
\bottomrule &  &  \\
\begin{tabularx}{\linewidth}{Xr}
		A \textbf{bold} value indicates higher accuracy, and \textit{\textbf{bold italics}} indicate a tie.  & \textbf{Unique Wins}
		\end{tabularx} & \textbf{9} & \textbf{23} \\
\cmidrule[0.4pt](lr){2-3} &  &  \\
\begin{tabularx}{\linewidth}{Xr} The test is a one-tailed binomial test to determine the probability that the strategy in the  
             rightmost column would achieve so many wins if wins and losses were equiprobable. & \textbf{Test Statistics} \end{tabularx} & \textbf{p-value: 0.01003} & \textbf{Confidence Interval:  0.56055 --- 1} \\

			\bottomrule
		\end{tabular}
	}
\end{table}

\subsection{Using weight accumulated at leaves to measure split evaluation intervals instead of number of instances seen}
\label{hat-getWeightSeen-nodeTime}

If examples at leaves are weighted, and split evaluation is done at \textit{intervals of a certain amount of weight seen}, weighted examples will cause more frequent splitting. As Poisson(1) is used for weighting, on the whole, the effects should be negligible. As with VFDT, we find (Table \ref{table5}) that there is no notable effect on the synthetic streams due to replacing a weight-based split evaluation timer (as in the original implementation) with an instance based one.

\begin{table}[!ht]
	
	\caption{\label{table5}Using getweightSeen instead of nodeTime: No measurable difference in prequential accuracy
	} 
	\scriptsize
	\centering
	\makebox[0.6\textwidth]{
		
		\begin{tabular}{m{12cm}?m{1.4cm}?m{1.4cm}}
			\toprule
			\multicolumn{1}{c?}{\textbf{Streams}} &  \multicolumn{1}{m{1.4cm}}{\textbf{HAT with nodeTime}} &
			\textbf{HAT with getWeightSeen}\\
			\midrule
			RecurrentConceptDriftStream -x 200000 -y 200000 -z 100 -s (AgrawalGenerator -f 2 -i 2) -d (AgrawalGenerator -f 3 -i 3) & \textit{\textbf{0.13645}} & \textit{\textbf{0.13645}} \\
RecurrentConceptDriftStream -x 200000 -y 200000 -z 100 -s (RandomTreeGenerator -r 1 -i 1) -d (RandomTreeGenerator -r 2 -i 2) & \textit{\textbf{0.10104}} & \textit{\textbf{0.10104}} \\
RecurrentConceptDriftStream -x 200000 -y 200000 -z 100 -s (SEAGenerator -f 2 -i 2) -d (SEAGenerator -f 3 -i 3) & \textit{\textbf{0.12579}} & \textit{\textbf{0.12579}} \\
RecurrentConceptDriftStream -x 200000 -y 200000 -z 100 -s (STAGGERGenerator -i 2 -f 2) -d (STAGGERGenerator -i 3 -f 3) & \textit{\textbf{0.00215}} & \textit{\textbf{0.00215}} \\
HyperplaneGenerator -k 10 -t 0.0001 -i 2 & \textit{\textbf{0.10801}} & \textit{\textbf{0.10801}} \\
HyperplaneGenerator -k 10 -t 0.001 -i 2 & \textit{\textbf{0.11934}} & \textit{\textbf{0.11934}} \\
HyperplaneGenerator -k 10 -t 0.01 -i 2 & \textit{\textbf{0.1198}} & \textit{\textbf{0.1198}} \\
HyperplaneGenerator -k 5 -t 0.0001 -i 2 & \textit{\textbf{0.10588}} & \textit{\textbf{0.10588}} \\
HyperplaneGenerator -k 5 -t 0.001 -i 2 & \textit{\textbf{0.11254}} & \textit{\textbf{0.11254}} \\
HyperplaneGenerator -k 5 -t 0.01 -i 2 & \textit{\textbf{0.10767}} & \textit{\textbf{0.10767}} \\
LEDGeneratorDrift -d 1 -i 2 & \textit{\textbf{0.26137}} & \textit{\textbf{0.26137}} \\
LEDGeneratorDrift -d 3 -i 2 & \textit{\textbf{0.26137}} & \textit{\textbf{0.26137}} \\
LEDGeneratorDrift -d 5 -i 2 & \textit{\textbf{0.26137}} & \textit{\textbf{0.26137}} \\
LEDGeneratorDrift -d 7 -i 2 & \textit{\textbf{0.26137}} & \textit{\textbf{0.26137}} \\
RandomRBFGeneratorDrift -s 0.0001 -k 10 -i 2 -r 2 & \textit{\textbf{0.13508}} & \textit{\textbf{0.13508}} \\
RandomRBFGeneratorDrift -s 0.0001 -k 50 -i 2 -r 2 & \textit{\textbf{0.19549}} & \textit{\textbf{0.19549}} \\
RandomRBFGeneratorDrift -s 0.001 -k 10 -i 2 -r 2 & \textit{\textbf{0.16636}} & \textit{\textbf{0.16636}} \\
RandomRBFGeneratorDrift -s 0.001 -k 50 -i 2 -r 2 & \textit{\textbf{0.33926}} & \textit{\textbf{0.33926}} \\
WaveformGeneratorDrift -d 1 -i 2 -n & \textit{\textbf{0.16279}} & \textit{\textbf{0.16279}} \\
WaveformGeneratorDrift -d 3 -i 2 -n & \textit{\textbf{0.16279}} & \textit{\textbf{0.16279}} \\
WaveformGeneratorDrift -d 5 -i 2 -n & \textit{\textbf{0.16279}} & \textit{\textbf{0.16279}} \\
WaveformGeneratorDrift -d 7 -i 2 -n & \textit{\textbf{0.16279}} & \textit{\textbf{0.16279}} \\
AbruptDriftGenerator -c  -o 1.0 -z 2 -n 2 -v 2 -r 2 -b 200000 -d Recurrent & \textit{\textbf{0.00185}} & \textit{\textbf{0.00185}} \\
AbruptDriftGenerator -c  -o 1.0 -z 3 -n 2 -v 2 -r 2 -b 200000 -d Recurrent & \textit{\textbf{0.0016}} & \textit{\textbf{0.0016}} \\
AbruptDriftGenerator -c  -o 1.0 -z 3 -n 3 -v 2 -r 2 -b 200000 -d Recurrent & \textit{\textbf{0.00252}} & \textit{\textbf{0.00252}} \\
AbruptDriftGenerator -c  -o 1.0 -z 3 -n 3 -v 3 -r 2 -b 200000 -d Recurrent & \textit{\textbf{0.00425}} & \textit{\textbf{0.00425}} \\
AbruptDriftGenerator -c  -o 1.0 -z 3 -n 3 -v 4 -r 2 -b 200000 -d Recurrent & \textit{\textbf{0.01212}} & \textit{\textbf{0.01212}} \\
AbruptDriftGenerator -c  -o 1.0 -z 3 -n 3 -v 5 -r 2 -b 200000 -d Recurrent & \textit{\textbf{0.01687}} & \textit{\textbf{0.01687}} \\
AbruptDriftGenerator -c  -o 1.0 -z 4 -n 2 -v 2 -r 2 -b 200000 -d Recurrent & \textit{\textbf{0.00177}} & \textit{\textbf{0.00177}} \\
AbruptDriftGenerator -c  -o 1.0 -z 4 -n 4 -v 4 -r 2 -b 200000 -d Recurrent & \textit{\textbf{0.06662}} & \textit{\textbf{0.06662}} \\
AbruptDriftGenerator -c  -o 1.0 -z 5 -n 2 -v 2 -r 2 -b 200000 -d Recurrent & \textit{\textbf{0.00177}} & \textit{\textbf{0.00177}} \\
AbruptDriftGenerator -c  -o 1.0 -z 5 -n 5 -v 5 -r 2 -b 200000 -d Recurrent & \textit{\textbf{0.34892}} & \textit{\textbf{0.34892}} \\

			\bottomrule
		\end{tabular}
	}
\end{table}

\subsection{Averaged Information Gain}
\label{hat-infogainavg}

According to the original specification, Information Gain needs to be averaged across all timesteps, as explained in Section \ref{ht-infogain}. Table \ref{table7} shows us that the HAT approximation strategy of using the latest computed infogain divided by the number of instances results in a net performance gain in prequential accuracy. Instantaneous Information Gain should be higher than a smoothed average if there is a trend favoring an attribute---which is particularly relevant in drifting settings.

\begin{table}[!ht]
	
	\caption{\label{table7}Averaging Information Gain
	} 
	\scriptsize
	\centering
	\makebox[0.6\textwidth]{
		
		\begin{tabular}{m{12cm}?m{1.4cm}?m{1.4cm}}
			\toprule
			\multicolumn{1}{c?}{\textbf{Streams}} &  \multicolumn{1}{m{1.4cm}}{\textbf{HAT}} &
			\textbf{HAT with averaged infogain}\\
			\midrule
			RecurrentConceptDriftStream -x 200000 -y 200000 -z 100 -s (AgrawalGenerator -f 2 -i 2) -d (AgrawalGenerator -f 3 -i 3) & \textbf{0.13645} & 0.13871 \\
RecurrentConceptDriftStream -x 200000 -y 200000 -z 100 -s (RandomTreeGenerator -r 1 -i 1) -d (RandomTreeGenerator -r 2 -i 2) & 0.10104 & \textbf{0.10079} \\
RecurrentConceptDriftStream -x 200000 -y 200000 -z 100 -s (SEAGenerator -f 2 -i 2) -d (SEAGenerator -f 3 -i 3) & 0.12579 & \textbf{0.12578} \\
RecurrentConceptDriftStream -x 200000 -y 200000 -z 100 -s (STAGGERGenerator -i 2 -f 2) -d (STAGGERGenerator -i 3 -f 3) & \textit{\textbf{0.00215}} & \textit{\textbf{0.00215}} \\
HyperplaneGenerator -k 10 -t 0.0001 -i 2 & \textbf{0.10801} & 0.10832 \\
HyperplaneGenerator -k 10 -t 0.001 -i 2 & 0.11934 & \textbf{0.11929} \\
HyperplaneGenerator -k 10 -t 0.01 -i 2 & 0.1198 & \textbf{0.11979} \\
HyperplaneGenerator -k 5 -t 0.0001 -i 2 & \textbf{0.10588} & 0.1063 \\
HyperplaneGenerator -k 5 -t 0.001 -i 2 & 0.11254 & \textbf{0.11148} \\
HyperplaneGenerator -k 5 -t 0.01 -i 2 & \textbf{0.10767} & 0.1083 \\
LEDGeneratorDrift -d 1 -i 2 & 0.26137 & \textbf{0.26121} \\
LEDGeneratorDrift -d 3 -i 2 & 0.26137 & \textbf{0.26121} \\
LEDGeneratorDrift -d 5 -i 2 & 0.26137 & \textbf{0.26121} \\
LEDGeneratorDrift -d 7 -i 2 & 0.26137 & \textbf{0.26121} \\
RandomRBFGeneratorDrift -s 0.0001 -k 10 -i 2 -r 2 & \textbf{0.13508} & 0.1372 \\
RandomRBFGeneratorDrift -s 0.0001 -k 50 -i 2 -r 2 & 0.19549 & \textbf{0.19518} \\
RandomRBFGeneratorDrift -s 0.001 -k 10 -i 2 -r 2 & 0.16636 & \textbf{0.16541} \\
RandomRBFGeneratorDrift -s 0.001 -k 50 -i 2 -r 2 & \textbf{0.33926} & 0.33939 \\
WaveformGeneratorDrift -d 1 -i 2 -n & 0.16279 & \textbf{0.1625} \\
WaveformGeneratorDrift -d 3 -i 2 -n & 0.16279 & \textbf{0.1625} \\
WaveformGeneratorDrift -d 5 -i 2 -n & 0.16279 & \textbf{0.1625} \\
WaveformGeneratorDrift -d 7 -i 2 -n & 0.16279 & \textbf{0.1625} \\
AbruptDriftGenerator -c  -o 1.0 -z 2 -n 2 -v 2 -r 2 -b 200000 -d Recurrent & 0.00185 & \textbf{0.00183} \\
AbruptDriftGenerator -c  -o 1.0 -z 3 -n 2 -v 2 -r 2 -b 200000 -d Recurrent & \textbf{0.0016} & 0.00161 \\
AbruptDriftGenerator -c  -o 1.0 -z 3 -n 3 -v 2 -r 2 -b 200000 -d Recurrent & \textbf{0.00252} & 0.0028 \\
AbruptDriftGenerator -c  -o 1.0 -z 3 -n 3 -v 3 -r 2 -b 200000 -d Recurrent & 0.00425 & \textbf{0.00421} \\
AbruptDriftGenerator -c  -o 1.0 -z 3 -n 3 -v 4 -r 2 -b 200000 -d Recurrent & 0.01212 & \textbf{0.0118} \\
AbruptDriftGenerator -c  -o 1.0 -z 3 -n 3 -v 5 -r 2 -b 200000 -d Recurrent & 0.01687 & \textbf{0.01646} \\
AbruptDriftGenerator -c  -o 1.0 -z 4 -n 2 -v 2 -r 2 -b 200000 -d Recurrent & 0.00177 & \textbf{0.00176} \\
AbruptDriftGenerator -c  -o 1.0 -z 4 -n 4 -v 4 -r 2 -b 200000 -d Recurrent & 0.06662 & \textbf{0.06559} \\
AbruptDriftGenerator -c  -o 1.0 -z 5 -n 2 -v 2 -r 2 -b 200000 -d Recurrent & 0.00177 & \textbf{0.00173} \\
AbruptDriftGenerator -c  -o 1.0 -z 5 -n 5 -v 5 -r 2 -b 200000 -d Recurrent & 0.34892 & \textbf{0.34081} \\
\bottomrule &  &  \\
\begin{tabularx}{\linewidth}{Xr}
		A \textbf{bold} value indicates higher accuracy, and \textit{\textbf{bold italics}} indicate a tie.  & \textbf{Unique Wins}
		\end{tabularx} & \textbf{8} & \textbf{23} \\
\cmidrule[0.4pt](lr){2-3} &  &  \\
\begin{tabularx}{\linewidth}{Xr} The test is a one-tailed binomial test to determine the probability that the strategy in the  
             rightmost column would achieve so many wins if wins and losses were equiprobable. & \textbf{Test Statistics} \end{tabularx} & \textbf{p-value: 0.00534} & \textbf{Confidence Interval:  0.58234 --- 1} \\

			\bottomrule
		\end{tabular}
	}
\end{table}

\subsection{Split replacement behaviors---1}
\label{hat-rootReplacement}

The original HAT exhibits some split replacement behaviors that appear to be detrimental to its performance under certain conditions. When a root level alternate leaf node $l_{RA}$ is ready to split, it replaces the root node $Root$. Note that replacement shouldn't be occuring unless the alternate has better accuracy!

Table \ref{table8} studies the replacement of the root by a root level alternate when the alternate splits, with all of the VFDT unspecified features enabled. It is clear that this behavior leads to a significant drop in performance. The explanation could lie in the fact that all of the knowledge in the form of subtree structure accumulated under the root is prematurely lost.

\begin{table}[!ht]
	
	\caption{\label{table8}When an root alternate splits, it replaces the root
	} 
	\scriptsize
	\centering
	\makebox[0.6\textwidth]{
		
		\begin{tabular}{m{12cm}?m{1.4cm}?m{1.4cm}}
			\toprule
			\multicolumn{1}{c?}{\textbf{Streams}} &  \multicolumn{1}{m{1.4cm}}{\textbf{HAT with VFDT unspecified features}} &
			\textbf{...with added root substitution when root alternate splits}\\
			\midrule
			RecurrentConceptDriftStream -x 200000 -y 200000 -z 100 -s (AgrawalGenerator -f 2 -i 2) -d (AgrawalGenerator -f 3 -i 3) & \textbf{0.11303} & 0.11828 \\
RecurrentConceptDriftStream -x 200000 -y 200000 -z 100 -s (RandomTreeGenerator -r 1 -i 1) -d (RandomTreeGenerator -r 2 -i 2) & \textit{\textbf{0.09855}} & \textit{\textbf{0.09855}} \\
RecurrentConceptDriftStream -x 200000 -y 200000 -z 100 -s (SEAGenerator -f 2 -i 2) -d (SEAGenerator -f 3 -i 3) & \textit{\textbf{0.11203}} & \textit{\textbf{0.11203}} \\
RecurrentConceptDriftStream -x 200000 -y 200000 -z 100 -s (STAGGERGenerator -i 2 -f 2) -d (STAGGERGenerator -i 3 -f 3) & 0.00213 & \textbf{0.00177} \\
HyperplaneGenerator -k 10 -t 0.0001 -i 2 & \textbf{0.10882} & 0.10884 \\
HyperplaneGenerator -k 10 -t 0.001 -i 2 & \textbf{0.11838} & 0.11855 \\
HyperplaneGenerator -k 10 -t 0.01 -i 2 & \textbf{0.11943} & 0.11982 \\
HyperplaneGenerator -k 5 -t 0.0001 -i 2 & \textbf{0.10675} & 0.10696 \\
HyperplaneGenerator -k 5 -t 0.001 -i 2 & \textbf{0.11007} & 0.11033 \\
HyperplaneGenerator -k 5 -t 0.01 -i 2 & \textbf{0.10528} & 0.10551 \\
LEDGeneratorDrift -d 1 -i 2 & \textit{\textbf{0.26121}} & \textit{\textbf{0.26121}} \\
LEDGeneratorDrift -d 3 -i 2 & \textit{\textbf{0.26121}} & \textit{\textbf{0.26121}} \\
LEDGeneratorDrift -d 5 -i 2 & \textit{\textbf{0.26121}} & \textit{\textbf{0.26121}} \\
LEDGeneratorDrift -d 7 -i 2 & \textit{\textbf{0.26121}} & \textit{\textbf{0.26121}} \\
RandomRBFGeneratorDrift -s 0.0001 -k 10 -i 2 -r 2 & \textit{\textbf{0.11975}} & \textit{\textbf{0.11975}} \\
RandomRBFGeneratorDrift -s 0.0001 -k 50 -i 2 -r 2 & 0.19195 & \textbf{0.19142} \\
RandomRBFGeneratorDrift -s 0.001 -k 10 -i 2 -r 2 & \textit{\textbf{0.15236}} & \textit{\textbf{0.15236}} \\
RandomRBFGeneratorDrift -s 0.001 -k 50 -i 2 -r 2 & 0.33901 & \textbf{0.3385} \\
WaveformGeneratorDrift -d 1 -i 2 -n & \textbf{0.16349} & 0.16367 \\
WaveformGeneratorDrift -d 3 -i 2 -n & \textbf{0.16349} & 0.16367 \\
WaveformGeneratorDrift -d 5 -i 2 -n & \textbf{0.16349} & 0.16367 \\
WaveformGeneratorDrift -d 7 -i 2 -n & \textbf{0.16349} & 0.16367 \\
AbruptDriftGenerator -c  -o 1.0 -z 2 -n 2 -v 2 -r 2 -b 200000 -d Recurrent & \textbf{0.00087} & 0.00093 \\
AbruptDriftGenerator -c  -o 1.0 -z 3 -n 2 -v 2 -r 2 -b 200000 -d Recurrent & \textbf{0.00083} & 0.00097 \\
AbruptDriftGenerator -c  -o 1.0 -z 3 -n 3 -v 2 -r 2 -b 200000 -d Recurrent & 0.00161 & \textbf{0.00152} \\
AbruptDriftGenerator -c  -o 1.0 -z 3 -n 3 -v 3 -r 2 -b 200000 -d Recurrent & \textbf{0.00412} & 0.00416 \\
AbruptDriftGenerator -c  -o 1.0 -z 3 -n 3 -v 4 -r 2 -b 200000 -d Recurrent & 0.01182 & \textbf{0.0117} \\
AbruptDriftGenerator -c  -o 1.0 -z 3 -n 3 -v 5 -r 2 -b 200000 -d Recurrent & \textbf{0.0164} & 0.01642 \\
AbruptDriftGenerator -c  -o 1.0 -z 4 -n 2 -v 2 -r 2 -b 200000 -d Recurrent & \textit{\textbf{0.00101}} & \textit{\textbf{0.00101}} \\
AbruptDriftGenerator -c  -o 1.0 -z 4 -n 4 -v 4 -r 2 -b 200000 -d Recurrent & \textit{\textbf{0.06557}} & \textit{\textbf{0.06557}} \\
AbruptDriftGenerator -c  -o 1.0 -z 5 -n 2 -v 2 -r 2 -b 200000 -d Recurrent & 0.00104 & \textbf{0.00101} \\
AbruptDriftGenerator -c  -o 1.0 -z 5 -n 5 -v 5 -r 2 -b 200000 -d Recurrent & \textit{\textbf{0.34081}} & \textit{\textbf{0.34081}} \\
\bottomrule &  &  \\
\begin{tabularx}{\linewidth}{Xr}
		A \textbf{bold} value indicates higher accuracy, and \textit{\textbf{bold italics}} indicate a tie.  & \textbf{Unique Wins}
		\end{tabularx} & \textbf{15} & \textbf{6} \\
\cmidrule[0.4pt](lr){2-3} &  &  \\
\begin{tabularx}{\linewidth}{Xr} The test is a one-tailed binomial test to determine the probability that the strategy in the  
             rightmost column would achieve so many wins if wins and losses were equiprobable. & \textbf{Test Statistics} \end{tabularx} & \textbf{p-value: 0.9867} & \textbf{Confidence Interval:  0.13245 --- 1} \\

			\bottomrule
		\end{tabular}
	}
\end{table}

\subsection{Split replacement behaviors---2}
\label{hat-subtreeAlternateReplacement}

When a leaf alternate $l_{A}$ of a mainline node ${notRoot}_1$ that is not the root splits, it replaces the corresponding mainline subtree of ${notRoot}_1$---again, it is not meant to do so until it has sprouted a subtree with higher accuracy. We study these behaviors with respect to a HAT with all the VFDT unspecified behaviors enabled.

Table \ref{table9} studies the replacement of the mainline subtrees by alternates when the alternates split. This also turns out to be detrimental---lookahead in these cases is too rapid.

\begin{table}[!ht]
	
	\caption{\label{table9}When a non-root alternate splits, it replaces its corresponding mainline node
	} 
	\scriptsize
	\centering
	\makebox[0.6\textwidth]{
		
		\begin{tabular}{m{12cm}?m{1.4cm}?m{1.4cm}}
			\toprule
			\multicolumn{1}{c?}{\textbf{Streams}} &  \multicolumn{1}{m{1.4cm}}{\textbf{HAT with VFDT unspecified features}} &
			\textbf{...with substitution of mainline node when alternate splits (not root level)}\\
			\midrule
			RecurrentConceptDriftStream -x 200000 -y 200000 -z 100 -s (AgrawalGenerator -f 2 -i 2) -d (AgrawalGenerator -f 3 -i 3) & 0.11303 & \textbf{0.11167} \\
RecurrentConceptDriftStream -x 200000 -y 200000 -z 100 -s (RandomTreeGenerator -r 1 -i 1) -d (RandomTreeGenerator -r 2 -i 2) & \textbf{0.09855} & 0.1128 \\
RecurrentConceptDriftStream -x 200000 -y 200000 -z 100 -s (SEAGenerator -f 2 -i 2) -d (SEAGenerator -f 3 -i 3) & \textbf{0.11203} & 0.11727 \\
RecurrentConceptDriftStream -x 200000 -y 200000 -z 100 -s (STAGGERGenerator -i 2 -f 2) -d (STAGGERGenerator -i 3 -f 3) & 0.00213 & \textbf{0.00211} \\
HyperplaneGenerator -k 10 -t 0.0001 -i 2 & \textbf{0.10882} & 0.11003 \\
HyperplaneGenerator -k 10 -t 0.001 -i 2 & 0.11838 & \textbf{0.11828} \\
HyperplaneGenerator -k 10 -t 0.01 -i 2 & 0.11943 & \textbf{0.11889} \\
HyperplaneGenerator -k 5 -t 0.0001 -i 2 & \textbf{0.10675} & 0.10986 \\
HyperplaneGenerator -k 5 -t 0.001 -i 2 & 0.11007 & \textbf{0.11003} \\
HyperplaneGenerator -k 5 -t 0.01 -i 2 & 0.10528 & \textbf{0.10476} \\
LEDGeneratorDrift -d 1 -i 2 & \textbf{0.26121} & 0.26129 \\
LEDGeneratorDrift -d 3 -i 2 & \textbf{0.26121} & 0.26129 \\
LEDGeneratorDrift -d 5 -i 2 & \textbf{0.26121} & 0.26129 \\
LEDGeneratorDrift -d 7 -i 2 & \textbf{0.26121} & 0.26129 \\
RandomRBFGeneratorDrift -s 0.0001 -k 10 -i 2 -r 2 & \textbf{0.11975} & 0.12437 \\
RandomRBFGeneratorDrift -s 0.0001 -k 50 -i 2 -r 2 & 0.19195 & \textbf{0.19145} \\
RandomRBFGeneratorDrift -s 0.001 -k 10 -i 2 -r 2 & \textbf{0.15236} & 0.15928 \\
RandomRBFGeneratorDrift -s 0.001 -k 50 -i 2 -r 2 & 0.33901 & \textbf{0.33875} \\
WaveformGeneratorDrift -d 1 -i 2 -n & \textbf{0.16349} & 0.17113 \\
WaveformGeneratorDrift -d 3 -i 2 -n & \textbf{0.16349} & 0.17113 \\
WaveformGeneratorDrift -d 5 -i 2 -n & \textbf{0.16349} & 0.17113 \\
WaveformGeneratorDrift -d 7 -i 2 -n & \textbf{0.16349} & 0.17113 \\
AbruptDriftGenerator -c  -o 1.0 -z 2 -n 2 -v 2 -r 2 -b 200000 -d Recurrent & 0.00087 & \textbf{0.00085} \\
AbruptDriftGenerator -c  -o 1.0 -z 3 -n 2 -v 2 -r 2 -b 200000 -d Recurrent & \textbf{0.00083} & 0.00084 \\
AbruptDriftGenerator -c  -o 1.0 -z 3 -n 3 -v 2 -r 2 -b 200000 -d Recurrent & \textbf{0.00161} & 0.00164 \\
AbruptDriftGenerator -c  -o 1.0 -z 3 -n 3 -v 3 -r 2 -b 200000 -d Recurrent & 0.00412 & \textbf{0.00405} \\
AbruptDriftGenerator -c  -o 1.0 -z 3 -n 3 -v 4 -r 2 -b 200000 -d Recurrent & \textbf{0.01182} & 0.01201 \\
AbruptDriftGenerator -c  -o 1.0 -z 3 -n 3 -v 5 -r 2 -b 200000 -d Recurrent & \textbf{0.0164} & 0.01832 \\
AbruptDriftGenerator -c  -o 1.0 -z 4 -n 2 -v 2 -r 2 -b 200000 -d Recurrent & 0.00101 & \textbf{0.00098} \\
AbruptDriftGenerator -c  -o 1.0 -z 4 -n 4 -v 4 -r 2 -b 200000 -d Recurrent & \textbf{0.06557} & 0.0675 \\
AbruptDriftGenerator -c  -o 1.0 -z 5 -n 2 -v 2 -r 2 -b 200000 -d Recurrent & \textbf{0.00104} & 0.00106 \\
AbruptDriftGenerator -c  -o 1.0 -z 5 -n 5 -v 5 -r 2 -b 200000 -d Recurrent & \textbf{0.34081} & 0.34213 \\
\bottomrule &  &  \\
\begin{tabularx}{\linewidth}{Xr}
		A \textbf{bold} value indicates higher accuracy, and \textit{\textbf{bold italics}} indicate a tie.  & \textbf{Unique Wins}
		\end{tabularx} & \textbf{21} & \textbf{11} \\
\cmidrule[0.4pt](lr){2-3} &  &  \\
\begin{tabularx}{\linewidth}{Xr} The test is a one-tailed binomial test to determine the probability that the strategy in the  
             rightmost column would achieve so many wins if wins and losses were equiprobable. & \textbf{Test Statistics} \end{tabularx} & \textbf{p-value: 0.97495} & \textbf{Confidence Interval:  0.20618 --- 1} \\

			\bottomrule
		\end{tabular}
	}
\end{table}

\subsection{Split replacement behaviors---3}
\label{replacement}

Following on from Sections \ref{hat-rootReplacement} and \ref{hat-subtreeAlternateReplacement}, we should expect the combination of premature root replacement and premature non-root replacement should lead to an even larger drop in prequential accuracy. Table \ref{table10} studies the case with both replacement of the mainline subtrees by alternates when the alternates split, and of the root when a root alternate splits. Interestingly, combining these strategies leads to an improvement over each strategy alone as shown in Tables \ref{table8} and \ref{table9}! While it does not perform significantly better than our rewritten baseline HAT, this can tip results in studies that focus mainly on naive rankings of algorithm pergormance.

\begin{table}[!ht]
	
	\caption{\label{table10}When an root alternate splits, it replaces the root; similarly, when any alternate splits, it replaces it's corresponding mainline node
	} 
	\scriptsize
	\centering
	\makebox[0.6\textwidth]{
		
		\begin{tabular}{m{12cm}?m{1.4cm}?m{1.4cm}}
			\toprule
			\multicolumn{1}{c?}{\textbf{Streams}} &  \multicolumn{1}{m{1.4cm}}{\textbf{HAT}} &
			\textbf{HAT with premature replacement by alternates at all levels}\\
			\midrule
			RecurrentConceptDriftStream -x 200000 -y 200000 -z 100 -s (AgrawalGenerator -f 2 -i 2) -d (AgrawalGenerator -f 3 -i 3) & 0.13098 & \textbf{0.12793} \\
RecurrentConceptDriftStream -x 200000 -y 200000 -z 100 -s (RandomTreeGenerator -r 1 -i 1) -d (RandomTreeGenerator -r 2 -i 2) & \textbf{0.0994} & 0.12227 \\
RecurrentConceptDriftStream -x 200000 -y 200000 -z 100 -s (SEAGenerator -f 2 -i 2) -d (SEAGenerator -f 3 -i 3) & \textbf{0.11224} & 0.11853 \\
RecurrentConceptDriftStream -x 200000 -y 200000 -z 100 -s (STAGGERGenerator -i 2 -f 2) -d (STAGGERGenerator -i 3 -f 3) & \textbf{0.00187} & 0.00188 \\
HyperplaneGenerator -k 10 -t 0.0001 -i 2 & 0.11936 & \textbf{0.11644} \\
HyperplaneGenerator -k 10 -t 0.001 -i 2 & 0.13425 & \textbf{0.1241} \\
HyperplaneGenerator -k 10 -t 0.01 -i 2 & 0.13181 & \textbf{0.12509} \\
HyperplaneGenerator -k 5 -t 0.0001 -i 2 & 0.13868 & \textbf{0.11753} \\
HyperplaneGenerator -k 5 -t 0.001 -i 2 & \textbf{0.11149} & 0.11526 \\
HyperplaneGenerator -k 5 -t 0.01 -i 2 & \textbf{0.10533} & 0.10867 \\
LEDGeneratorDrift -d 1 -i 2 & \textbf{0.26118} & 0.26124 \\
LEDGeneratorDrift -d 3 -i 2 & \textbf{0.26118} & 0.26124 \\
LEDGeneratorDrift -d 5 -i 2 & \textbf{0.26118} & 0.26124 \\
LEDGeneratorDrift -d 7 -i 2 & \textbf{0.26118} & 0.26124 \\
RandomRBFGeneratorDrift -s 0.0001 -k 10 -i 2 -r 2 & 0.13354 & \textbf{0.12037} \\
RandomRBFGeneratorDrift -s 0.0001 -k 50 -i 2 -r 2 & 0.18126 & \textbf{0.18017} \\
RandomRBFGeneratorDrift -s 0.001 -k 10 -i 2 -r 2 & 0.19556 & \textbf{0.15522} \\
RandomRBFGeneratorDrift -s 0.001 -k 50 -i 2 -r 2 & \textbf{0.32677} & 0.32757 \\
WaveformGeneratorDrift -d 1 -i 2 -n & 0.17888 & \textbf{0.17118} \\
WaveformGeneratorDrift -d 3 -i 2 -n & 0.17888 & \textbf{0.17118} \\
WaveformGeneratorDrift -d 5 -i 2 -n & 0.17888 & \textbf{0.17118} \\
WaveformGeneratorDrift -d 7 -i 2 -n & 0.17887 & \textbf{0.17118} \\
AbruptDriftGenerator -c  -o 1.0 -z 2 -n 2 -v 2 -r 2 -b 200000 -d Recurrent & \textbf{0.00065} & 0.00068 \\
AbruptDriftGenerator -c  -o 1.0 -z 3 -n 2 -v 2 -r 2 -b 200000 -d Recurrent & 0.00068 & \textbf{0.00062} \\
AbruptDriftGenerator -c  -o 1.0 -z 3 -n 3 -v 2 -r 2 -b 200000 -d Recurrent & 0.00167 & \textbf{0.00122} \\
AbruptDriftGenerator -c  -o 1.0 -z 3 -n 3 -v 3 -r 2 -b 200000 -d Recurrent & 0.00642 & \textbf{0.00307} \\
AbruptDriftGenerator -c  -o 1.0 -z 3 -n 3 -v 4 -r 2 -b 200000 -d Recurrent & 0.01252 & \textbf{0.0093} \\
AbruptDriftGenerator -c  -o 1.0 -z 3 -n 3 -v 5 -r 2 -b 200000 -d Recurrent & \textbf{0.01473} & 0.01475 \\
AbruptDriftGenerator -c  -o 1.0 -z 4 -n 2 -v 2 -r 2 -b 200000 -d Recurrent & \textbf{0.00069} & 0.00073 \\
AbruptDriftGenerator -c  -o 1.0 -z 4 -n 4 -v 4 -r 2 -b 200000 -d Recurrent & \textbf{0.05099} & 0.05314 \\
AbruptDriftGenerator -c  -o 1.0 -z 5 -n 2 -v 2 -r 2 -b 200000 -d Recurrent & 0.00079 & \textbf{0.00078} \\
AbruptDriftGenerator -c  -o 1.0 -z 5 -n 5 -v 5 -r 2 -b 200000 -d Recurrent & \textbf{0.28127} & 0.28637 \\
\bottomrule &  &  \\
\begin{tabularx}{\linewidth}{Xr}
		A \textbf{bold} value indicates higher accuracy, and \textit{\textbf{bold italics}} indicate a tie.  & \textbf{Unique Wins}
		\end{tabularx} & \textbf{15} & \textbf{17} \\
\cmidrule[0.4pt](lr){2-3} &  &  \\
\begin{tabularx}{\linewidth}{Xr} The test is a one-tailed binomial test to determine the probability that the strategy in the  
             rightmost column would achieve so many wins if wins and losses were equiprobable. & \textbf{Test Statistics} \end{tabularx} & \textbf{p-value: 0.43003} & \textbf{Confidence Interval:  0.37339 --- 1} \\

			\bottomrule
		\end{tabular}
	}
\end{table}

\subsection{Effects from HoeffdingTree}
\label{hat-ht-effects}

Table \ref{table11} shows us that adding just the unspecified features from HoeffdingTree is beneficial for HAT, as it was found to be for VFDT. 

There are many more combinations and possibilities to study---given there are 9 options, a simple binomial sum gives us 19,863 possibilities to study. It is possible there are scenarios in which split replacement behaviors lead to an improvement under concept drift on the testbench.

\begin{table}[!ht]
	
	\caption{\label{table11}With and without unspecified VFDT features 
	} 
	\scriptsize
	\centering
	\makebox[0.6\textwidth]{
		
		\begin{tabular}{m{12cm}?m{1.4cm}?m{1.4cm}}
			\toprule
			\multicolumn{1}{c?}{\textbf{Streams}} &  \multicolumn{1}{m{1.4cm}}{\textbf{HAT}} &
			\textbf{HAT with VFDT unspecified features}\\
			\midrule
			RecurrentConceptDriftStream -x 200000 -y 200000 -z 100 -s (AgrawalGenerator -f 2 -i 2) -d (AgrawalGenerator -f 3 -i 3) & 0.13645 & \textbf{0.11303} \\
RecurrentConceptDriftStream -x 200000 -y 200000 -z 100 -s (RandomTreeGenerator -r 1 -i 1) -d (RandomTreeGenerator -r 2 -i 2) & 0.10104 & \textbf{0.09855} \\
RecurrentConceptDriftStream -x 200000 -y 200000 -z 100 -s (SEAGenerator -f 2 -i 2) -d (SEAGenerator -f 3 -i 3) & 0.12579 & \textbf{0.11203} \\
RecurrentConceptDriftStream -x 200000 -y 200000 -z 100 -s (STAGGERGenerator -i 2 -f 2) -d (STAGGERGenerator -i 3 -f 3) & 0.00215 & \textbf{0.00213} \\
HyperplaneGenerator -k 10 -t 0.0001 -i 2 & \textbf{0.10801} & 0.10882 \\
HyperplaneGenerator -k 10 -t 0.001 -i 2 & 0.11934 & \textbf{0.11838} \\
HyperplaneGenerator -k 10 -t 0.01 -i 2 & 0.1198 & \textbf{0.11943} \\
HyperplaneGenerator -k 5 -t 0.0001 -i 2 & \textbf{0.10588} & 0.10675 \\
HyperplaneGenerator -k 5 -t 0.001 -i 2 & 0.11254 & \textbf{0.11007} \\
HyperplaneGenerator -k 5 -t 0.01 -i 2 & 0.10767 & \textbf{0.10528} \\
LEDGeneratorDrift -d 1 -i 2 & 0.26137 & \textbf{0.26121} \\
LEDGeneratorDrift -d 3 -i 2 & 0.26137 & \textbf{0.26121} \\
LEDGeneratorDrift -d 5 -i 2 & 0.26137 & \textbf{0.26121} \\
LEDGeneratorDrift -d 7 -i 2 & 0.26137 & \textbf{0.26121} \\
RandomRBFGeneratorDrift -s 0.0001 -k 10 -i 2 -r 2 & 0.13508 & \textbf{0.11975} \\
RandomRBFGeneratorDrift -s 0.0001 -k 50 -i 2 -r 2 & 0.19549 & \textbf{0.19195} \\
RandomRBFGeneratorDrift -s 0.001 -k 10 -i 2 -r 2 & 0.16636 & \textbf{0.15236} \\
RandomRBFGeneratorDrift -s 0.001 -k 50 -i 2 -r 2 & 0.33926 & \textbf{0.33901} \\
WaveformGeneratorDrift -d 1 -i 2 -n & \textbf{0.16279} & 0.16349 \\
WaveformGeneratorDrift -d 3 -i 2 -n & \textbf{0.16279} & 0.16349 \\
WaveformGeneratorDrift -d 5 -i 2 -n & \textbf{0.16279} & 0.16349 \\
WaveformGeneratorDrift -d 7 -i 2 -n & \textbf{0.16279} & 0.16349 \\
AbruptDriftGenerator -c  -o 1.0 -z 2 -n 2 -v 2 -r 2 -b 200000 -d Recurrent & 0.00185 & \textbf{0.00087} \\
AbruptDriftGenerator -c  -o 1.0 -z 3 -n 2 -v 2 -r 2 -b 200000 -d Recurrent & 0.0016 & \textbf{0.00083} \\
AbruptDriftGenerator -c  -o 1.0 -z 3 -n 3 -v 2 -r 2 -b 200000 -d Recurrent & 0.00252 & \textbf{0.00161} \\
AbruptDriftGenerator -c  -o 1.0 -z 3 -n 3 -v 3 -r 2 -b 200000 -d Recurrent & 0.00425 & \textbf{0.00412} \\
AbruptDriftGenerator -c  -o 1.0 -z 3 -n 3 -v 4 -r 2 -b 200000 -d Recurrent & 0.01212 & \textbf{0.01182} \\
AbruptDriftGenerator -c  -o 1.0 -z 3 -n 3 -v 5 -r 2 -b 200000 -d Recurrent & 0.01687 & \textbf{0.0164} \\
AbruptDriftGenerator -c  -o 1.0 -z 4 -n 2 -v 2 -r 2 -b 200000 -d Recurrent & 0.00177 & \textbf{0.00101} \\
AbruptDriftGenerator -c  -o 1.0 -z 4 -n 4 -v 4 -r 2 -b 200000 -d Recurrent & 0.06662 & \textbf{0.06557} \\
AbruptDriftGenerator -c  -o 1.0 -z 5 -n 2 -v 2 -r 2 -b 200000 -d Recurrent & 0.00177 & \textbf{0.00104} \\
AbruptDriftGenerator -c  -o 1.0 -z 5 -n 5 -v 5 -r 2 -b 200000 -d Recurrent & 0.34892 & \textbf{0.34081} \\
\bottomrule &  &  \\
\begin{tabularx}{\linewidth}{Xr}
		A \textbf{bold} value indicates higher accuracy, and \textit{\textbf{bold italics}} indicate a tie.  & \textbf{Unique Wins}
		\end{tabularx} & \textbf{6} & \textbf{26} \\
\cmidrule[0.4pt](lr){2-3} &  &  \\
\begin{tabularx}{\linewidth}{Xr} The test is a one-tailed binomial test to determine the probability that the strategy in the  
             rightmost column would achieve so many wins if wins and losses were equiprobable. & \textbf{Test Statistics} \end{tabularx} & \textbf{p-value: 0.00027} & \textbf{Confidence Interval:  0.66313 --- 1} \\

			\bottomrule
		\end{tabular}
	}
\end{table}

	\section{Conclusions}
\label{conclusions}

It is instructive to understand why a particular implementation of a proposed strategy works in terms of both unspecified features and intended mechanisms. We find that both VFDT and HAT have unspecified features that interact in complex ways to determine algorithm performance under concept drift.

In studying HAT, we came to understand the effects due to VFDT; in particular, the significance of inadvertent amnesia both by design of Hoeffding Tree (the theoretical artifact VFDT implements) and by design of VFDT-MOA (due to resplitting on used attributes). Proceeding along this line of inquiry, we also found that both major VFDT implementations (VFML and MOA) do not average infogain. Given Hoeffding Bound itself might not be suitable to use in the current context \parencite{rutkowski2012decision}, we find the approximations in computing the Hoeffding Bound are generally effective and harmless... with the added benefit of better performance on drifting streams.

We find the VFDT strategies of erasing node statistics through resplitting and approximating infogains individually somewhat effective, but significantly so when combined in terms of evoking a good response to concept drift from both VFDT, and HAT which derives from VFDT. We also find that the unspecified strategy of allowing alternates to vote in HAT significantly improves performance on streams with concept drift, with other strategies such as weighting leaves or optimistically early replacement of subtrees turning out to be beneficial or detrimental. We note that there are potentially thousands of possible interactions that need to be studied in order to extricate the true breadth of performance characteristics of simple strategies such as VFDT and HAT, and the choice of testbench adds further complexity to the conclusions one can draw.

A standardised, rationalised testbench that accounts for data stream characteristics does not exist in spite of decades of development in the field; we hope to have made a start in offering such a testbench. 

In practice, we suggest that the node evisceration strategy proposed in Section \ref{ht-evisceration} is integrated into standard Hoeffding Tree (and consequently into HAT) when they are expected to encounter concept drift, and that not averaging information gain \ref{ht-infogain} should continue being used in place of averaging information gain given the lack of evidence for adverse effect. However, it might be instructive to look for specific cases in which an effect is observed due to not averaging information gain. We also recommend that HAT is used with a single alternate until the effects of using multiple alternates are studied in greater depth, and that split replacement behaviors are normalised in general use (without the unspecified behaviors in Sections \ref{hat-resplitting}, \ref{hat-rootReplacement}, and \ref{hat-subtreeAlternateReplacement}), but that the mixture of unspecified strategies in Section \ref{replacement} be studied in more detail. 

It would be of immense utility to distil unspecified features from other basic machine learning methods so we may further our understanding of how learning strategies work.

\printbibliography

@inproceedings{bifet2009adaptive,
  title={Adaptive learning from evolving data streams},
  author={Bifet, Albert and Gavald{\`a}, Ricard},
  booktitle={International Symposium on Intelligent Data Analysis},
  pages={249--260},
  year={2009},
  organization={Springer}
}

@inproceedings{bifet2007learning,
  title={Learning from time-changing data with adaptive windowing},
  author={Bifet, Albert and Gavalda, Ricard},
  booktitle={Proceedings of the 2007 SIAM International Conference on Data Mining},
  pages={443--448},
  year={2007},
  organization={SIAM}
}

@article{bifet2010moa,
  title={Moa: Massive online analysis},
  author={Bifet, Albert and Holmes, Geoff and Kirkby, Richard and Pfahringer, Bernhard},
  journal={Journal of Machine Learning Research},
  volume={11},
  number={May},
  pages={1601--1604},
  year={2010}
}

@article{webb2016characterizing,
  title={Characterizing concept drift},
  author={Webb, Geoffrey I and Hyde, Roy and Cao, Hong and Nguyen, Hai Long and Petitjean, Francois},
  journal={Data Mining and Knowledge Discovery},
  volume={30},
  number={4},
  pages={964--994},
  year={2016},
  publisher={Springer}
}

@article{Gama:eval_survey,
 author = {Gama, João and Sebastiao, Raquel and Rodrigues, Pedro Pereira},
 title = {On Evaluating Stream Learning Algorithms},
 journal = {Mach. Learn.},
 issue_date = {March     2013},
 volume = {90},
 number = {3},
 month = mar,
 year = {2013},
 issn = {0885-6125},
 pages = {317--346},
 numpages = {30},
 url = {http://dx.doi.org/10.1007/s10994-012-5320-9},
 doi = {10.1007/s10994-012-5320-9},
 acmid = {2441014},
 publisher = {Kluwer Academic Publishers},
 address = {Hingham, MA, USA},
}

@article{hoens2012learning,
	title={Learning from streaming data with concept drift and imbalance: an overview},
	author={Hoens, T Ryan and Polikar, Robi and Chawla, Nitesh V},
	journal={Progress in Artificial Intelligence},
	volume={1},
	number={1},
	pages={89--101},
	year={2012},
	publisher={Springer}
}

@article{grossberg1988nonlinear,
	title={Nonlinear neural networks: Principles, mechanisms, and architectures},
	author={Grossberg, Stephen},
	journal={Neural networks},
	volume={1},
	number={1},
	pages={17--61},
	year={1988},
	publisher={Elsevier}
}

@article{breiman1984classification,
  title={Classification and regression trees},
  author={Breiman, Leo},
  year={1984},
  publisher={Wadsworth International Group}
}

@article{rutkowski2012decision,
	title={Decision trees for mining data streams based on the McDiarmid's bound},
	author={Rutkowski, Leszek and Pietruczuk, Lena and Duda, Piotr and Jaworski, Maciej},
	journal={IEEE Transactions on Knowledge and Data Engineering},
	volume={25},
	number={6},
	pages={1272--1279},
	year={2012},
	publisher={IEEE}
}

@inproceedings{domingos2000mining,
  title={Mining high-speed data streams},
  author={Domingos, Pedro and Hulten, Geoff},
  booktitle={Proceedings of the sixth ACM SIGKDD international conference on Knowledge discovery and data mining},
  pages={71--80},
  year={2000},
  organization={ACM}
}

@inproceedings{oza05,
	author={Nikunj C. Oza},
	title={Online Bagging and Boosting},
	booktitle={International Conference on Systems, Man, and Cybernetics, Special Session on Ensemble Methods for Extreme Environments}, 
	publisher={Institute for Electrical and Electronics Engineers},
	address={New Jersey},
	editor={Mo Jamshidi},
	pages={2340-2345},
	month={October},
	bib2html_pubtype={Refereed Conference},
	bib2html_rescat={Ensemble Learning},
	year={2005}
}

@book{Quinlan:2031749,
      author        = "Quinlan, J Ross",
      title         = "{C4.5: programs for machine learning}",
      publisher     = "Morgan Kaufmann",
      address       = "San Mateo, CA",
      year          = "1992",
      url           = "http://cds.cern.ch/record/2031749",
}

@article{hoeffding1963probability,
  title={Probability inequalities for sums of bounded random variables},
  author={Hoeffding, Wassily},
  journal={Journal of the American statistical association},
  volume={58},
  number={301},
  pages={13--30},
  year={1963},
  publisher={Taylor \& Francis Group}
}

@inproceedings{schlimmer1986case,
	title={A case study of incremental concept induction},
	author={Schlimmer, Jeffrey C and Fisher, Douglas},
	booktitle={AAAI},
	volume={86},
	pages={496--501},
	year={1986}
}

@article{utgoff1989incremental,
	title={Incremental induction of decision trees},
	author={Utgoff, Paul E},
	journal={Machine learning},
	volume={4},
	number={2},
	pages={161--186},
	year={1989},
	publisher={Springer}
}

@inproceedings{Gehrke:1999:BDT:304182.304197,
	author = {Gehrke, Johannes and Ganti, Venkatesh and Ramakrishnan, Raghu and Loh, Wei-Yin},
	title = {BOAT---Optimistic Decision Tree Construction},
	booktitle = {Proceedings of the 1999 ACM SIGMOD International Conference on Management of Data},
	series = {SIGMOD '99},
	year = {1999},
	isbn = {1-58113-084-8},
	location = {Philadelphia, Pennsylvania, USA},
	pages = {169--180},
	numpages = {12},
	url = {http://doi.acm.org/10.1145/304182.304197},
	doi = {10.1145/304182.304197},
	acmid = {304197},
	publisher = {ACM},
	address = {New York, NY, USA},
}

@article{gehrke2000rainforest,
	title={RainForest—a framework for fast decision tree construction of large datasets},
	author={Gehrke, Johannes and Ramakrishnan, Raghu and Ganti, Venkatesh},
	journal={Data Mining and Knowledge Discovery},
	volume={4},
	number={2-3},
	pages={127--162},
	year={2000},
	publisher={Springer}
}

@article{hunt1962concept,
	title={Concept learning: An information processing problem.},
	author={Hunt, Earl B},
	year={1962},
	publisher={John Wiley \& Sons Inc}
}

@book{earl1966experiments,
	title={Experiments in Induction},
	author={Earl Busby Hunt and Janet Marin and Philip James Stone},
	lccn={lc65026400},
	url={https://books.google.com.au/books?id=60NDAAAAIAAJ},
	year={1966},
	publisher={Academic Press}
}

@ARTICLE{Quinlan86inductionof,
	author = {J. R. Quinlan},
	title = {Induction of Decision Trees},
	journal = {MACH. LEARN},
	year = {1986},
	volume = {1},
	pages = {81--106}
}

@article{quinlan1979discovering,
	title={Discovering rules by induction from large collections of examples},
	author={Quinlan, J Ross},
	journal={Expert systems in the micro electronics age},
	year={1979},
	publisher={Edinburgh University Press}
}

@incollection{quinlan1983learning,
	title={Learning efficient classification procedures and their application to chess end games},
	author={Quinlan, J Ross},
	booktitle={Machine learning},
	pages={463--482},
	year={1983},
	publisher={Springer}
}
\end{document}